%% file: iclr2027_conference.tex
\documentclass{article} 
\usepackage{iclr2027_conference,times}

\input{math_commands.tex}

\usepackage{url}
\usepackage{graphicx}
\usepackage[utf8]{inputenc}

\usepackage{booktabs}
\usepackage{tabularx}
\usepackage{array}
\usepackage{float}
\usepackage{longtable}
\usepackage{hyperref}

\title{Hyperbolic Manifold Constrained Tabular Neural Network}

\iclrfinalcopy

\author{Tian Li \& Huizhi Liang \& Lucy Robinson \& Varun Ojha \\
School of Computing\\
Newcastle University\\
Newcastle upon Tyne, NE1 7RU, UK \\
\texttt{\{t.li56,Huizhi.Liang,Lucy.Robinson2,Varun.Ojha\}@newcastle.ac.uk} 
}

\begin{document}

\maketitle
\lhead{Under review as aconferencepaper at ICLR2027}

\begin{abstract}

Tabular prediction is central to a wide range of real-world applications.
Tabular data typically contain heterogeneous features as well as rich and complex relational information that can imply a latent structural manifold.
Hyperbolic geometry can help capture complex structural relations in data. 
However, most existing tabular prediction models are constructed and optimized in Euclidean space.
How to incorporate hyperbolic geometry into supervised tabular learning remains underexplored.
We propose \textbf{HTNN}, a supervised hyperbolic manifold constrained tabular neural network for tabular prediction.
HTNN consists of a hyperbolic feature-value representation layer for heterogeneous categorical and numerical features, followed by a conventional MLP predictor. HTNN employs a \emph{geometry-aware training} and \emph{geometry-free inference} optimization framework.  
The \emph{geometry-aware training} allows hyperbolic geometry to shape the latent representation learning of heterogeneous feature values. After training, the latent hyperbolic representations can be converted into ordinary Euclidean space for efficient \emph{geometry-free inference}.
We conducted extensive experiments on the TALENT benchmark.
HTNN ranks first among 36 methods on 200 classification datasets and third among 34 methods on 100 regression datasets. 
Experimental results show that the proposed hyperbolic manifold constrained tabular neural network is effective.

\end{abstract}

\section{Introduction}
\label{sec:intro}
Tabular prediction supports decision making across healthcare, finance, energy, and many other real-world domains~\citep{borisov2022deep}.
Tabular data typically contain heterogeneous categorical (e.g., gender) and numerical (e.g., age) features, together with rich and complex relational information that can imply latent manifold structure~\citep{choi2026tabgeoflow,he2025crafting}.
Unlike graphs or other explicitly structured inputs, tabular data provide no predefined connectivity or known hierarchy for these relationships, leaving them to be inferred from the data.
Existing studies have sought to model these implicit relationships through advances in feature encoding, interaction modeling, and training strategies~\citep{mlp_plr,fttransformer,realmlp,tabm}.
However, how to better learn latent representations of heterogeneous features and incorporate their implied manifold structure remains an open research question for tabular prediction models.

Hyperbolic geometry can help capture complex structural relationships in data.
Under negative curvature, hyperbolic space expands rapidly with distance from the origin, providing a distinctive geometry for organizing such relationships.
This property has motivated successful hyperbolic representation learning across hierarchies, graphs, vision, and multimodal learning~\citep{nickel2017poincare,chami2019hyperbolic,dai2021hyperbolic,desai2023hyperbolic}.
However, most existing tabular prediction models focus on Euclidean space~\citep{realmlp,modernnca,xrfm,tabpfn3}.
The use of hyperbolic geometry in supervised tabular learning remains underexplored.
In particular, it remains unclear how heterogeneous features should be represented within a unified hyperbolic space and how the resulting manifold geometry shapes their learning.

In this paper, we propose \textbf{HTNN}, a supervised hyperbolic manifold constrained tabular neural network for tabular prediction.
HTNN introduces a hyperbolic feature-value (HFV) representation layer with dedicated categorical and numerical pathways.
Each categorical value is assigned an individual learnable HFV representation.
Numerical features, however, do not provide a fixed finite vocabulary for direct representation lookup.
Each numerical value is therefore represented through an input-dependent composition of a small set of feature-specific HFV representations.
Both pathways use tangent readouts of their HFV representations, which are concatenated and processed by a conventional MLP predictor.
During \emph{geometry-aware training}, Riemannian optimization allows negative curvature to shape representation updates, producing radial adaptability with transverse stabilization.
After training, categorical tangent readouts are stored as ordinary lookup vectors, while the numerical pathway collapses into affine maps. 
This preserves the original predictions while removing all manifold operations, enabling \emph{geometry-free inference}.

We evaluate HTNN at two complementary scales.
The full 300-dataset TALENT benchmark~\citep{talent} provides the main predictive comparison, while the 45-dataset TALENT-tiny suite is used for controlled studies of optimization behavior, model design, and computational efficiency.
Our contributions are summarized as follows:
\begin{itemize}
\item We propose \textbf{HTNN}, a supervised hyperbolic manifold constrained tabular neural network that combines an HFV representation layer for heterogeneous categorical and numerical features with a conventional MLP predictor.
\item We introduce a \emph{geometry-aware training and geometry-free inference} framework: HFV representations are optimized on a hyperbolic manifold during training, and the trained HFV representation layer is converted into ordinary Euclidean lookup and affine operations for efficient inference.
\item We conducted extensive comparison experiments. On the full TALENT benchmark, HTNN ranks first among 36 methods on 200 classification datasets and third among 34 methods on 100 regression datasets.
\end{itemize}

\section{Related Work}
\label{sec:related_work}

\noindent\textbf{Feature representations in tabular neural networks.}
Neural tabular models have explored a wide range of architectures and input representations.
MLP-style and feature-attentive models learn from embedded feature values, while TabTransformer and FT-Transformer use attention to model feature interactions~\citep{tabmlp,tabnet,tabtransformer,fttransformer}.
Numerical embedding methods enrich scalar inputs through piecewise-linear or periodic representations, and RealMLP and TabM improve training strategies, default configurations, and parameter-efficient ensembling~\citep{mlp_plr,realmlp,tabm}.
Other approaches incorporate sample neighborhoods through retrieval or exploit cross-dataset context through pretrained foundation models~\citep{tabr,modernnca,tabpfnv2,tabicl}.
Despite this architectural diversity, most existing tabular models are constructed and optimized in Euclidean space.
Consequently, the role of non-Euclidean geometry in supervised tabular learning remains underexplored.

\noindent\textbf{Hyperbolic representation learning.}
Hyperbolic geometry has been widely used to represent hierarchical and relational information.
Poincar\'e embeddings exploit negative curvature to embed explicit hierarchies with low distortion~\citep{nickel2017poincare}, while subsequent work extends hyperbolic learning to graph neural networks and visual or multimodal representations~\citep{chami2019hyperbolic,dai2021hyperbolic,lensink2022fully,desai2023hyperbolic}.
HyperDT constructs decision trees for samples that are already represented in hyperbolic space~\citep{hyperdt}.
These methods generally begin with an explicit relational structure or an existing sample-level representation.
Without either, how hyperbolic geometry should be incorporated into tabular neural networks remains an open question.

\noindent\textbf{Training parameterization and Riemannian optimization.}
Equivalent predictors can follow different optimization trajectories under different trainable parameterizations~\citep{gunasekar2017implicit,arora2018optimization}.
Riemannian optimization enables manifold-valued parameters to be trained while respecting their geometric constraints, with RiemannianAdam extending adaptive optimization to such parameters~\citep{becigneul2018riemannian}.
However, the effects of manifold constraints on feature-value learning and supervised tabular prediction remain insufficiently characterized.

\section{Methodology}
\label{sec:method_overview}

\begin{figure}[H]
	\vskip -10pt
    \centering
    \includegraphics[width=\textwidth]{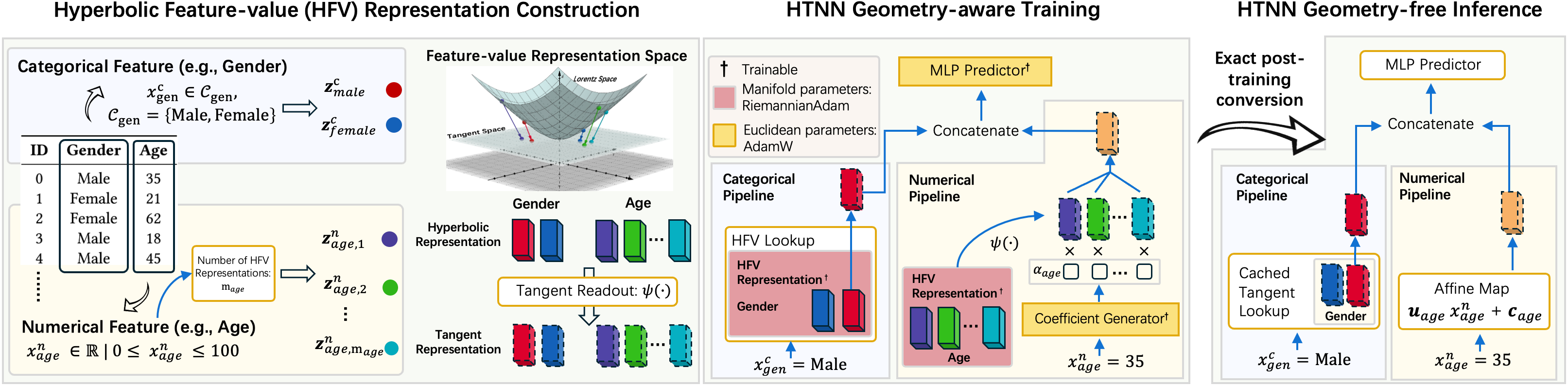}
    \vskip -10pt
    \caption{
Overview of HTNN. Categorical values select individual HFV representations, whereas the feature-wise capacity \(m_j\) determines how many HFV representations are assigned to numerical feature \(j\). Input-dependent coefficients combine their tangent readouts (left). 
HTNN jointly optimizes HFV representations and Euclidean parameters during \emph{geometry-aware training} (center). 
Exact post-training conversion caches categorical tangent readouts and collapses each numerical composition into an affine map for \emph{geometry-free inference} without changing predictions (right).
    }
    \vskip -10pt
    \label{fig:htnn_arch}
\end{figure}

HTNN is a supervised tabular prediction model that uses hyperbolic representation learning to construct feature-value representations for heterogeneous inputs.
An input $\mathbf{x}$ consists of categorical features $\{x_s^{\mathrm{c}}\in\mathcal{C}_s\}_{s\in\mathcal{J}_{\mathrm{c}}}$ and numerical features $\{x_j^{\mathrm{n}}\in\mathbb{R}\}_{j\in\mathcal{J}_{\mathrm{n}}}$, where $\mathcal{J}_{\mathrm{c}}$ and $\mathcal{J}_{\mathrm{n}}$ denote the two feature types and $\mathcal{C}_s$ is the value set of categorical feature $s$.
HTNN learns hyperbolic feature-value (HFV) representations for both feature types.
Each categorical value maps to an HFV representation, whereas each numerical value determines an input-dependent combination of the tangent readouts of a small set of feature-specific HFV representations.
Both pathways produce vectors in a shared tangent space for a conventional MLP predictor. 
Figure~\ref{fig:htnn_arch} summarizes the complete pipeline: the HFV representation construction, \emph{geometry-aware training}, and \emph{geometry-free inference}.

\subsection{Categorical HFV Representations}
\label{sec:hyperbolic_representations}
Both feature pathways use learnable HFV representations on a common Lorentz manifold.
The Lorentz model represents a \(d\)-dimensional hyperbolic space as a hyperboloid in the ambient space \(\mathbb{R}^{d+1}\), with one time-like coordinate and \(d\) spatial coordinates.
For $\mathbf{z},\mathbf{z}'\in\mathbb{R}^{d+1}$, the Lorentz inner product is
\vspace{-10pt}
\begin{equation}
\small
\left\langle \mathbf{z},\mathbf{z}'\right\rangle_{\mathrm{L}}
=
-z_0z'_0+\sum_{\ell=1}^{d}z_\ell z'_\ell .
\label{eq:lorentz_inner_product}
\end{equation}
Given a scale parameter $\kappa>0$, the Lorentz manifold is the upper sheet of this hyperboloid:
\begin{equation}
\small
\mathbb{H}_{\kappa}^{d}
=
\left\{
\mathbf{z}\in\mathbb{R}^{d+1}
:
\left\langle\mathbf{z},\mathbf{z}\right\rangle_{\mathrm{L}}
=-\kappa,\;
z_0>0
\right\}.
\label{eq:lorentz_manifold}
\end{equation}
This manifold has sectional curvature $-1/\kappa$ and origin $\mathbf{o}=(\sqrt{\kappa},0,\ldots,0)$.
To interface its HFV representations with a conventional neural predictor, we identify the tangent space at $\mathbf{o}$ with $\mathbb{R}^{d}$ through its spatial coordinates and define the tangent readout
\begin{equation}
\small
\psi(\mathbf{z})
=
\left[
\log_{\mathbf{o}}^{\mathbb{H}}(\mathbf{z})
\right]_{1:d}
\in\mathbb{R}^{d},
\label{eq:tangent_readout}
\end{equation}
where $\log_{\mathbf{o}}^{\mathbb{H}}$ is the Lorentz logarithmic map and $[\cdot]_{1:d}$ selects the spatial coordinates.
Thus, each HFV representation remains constrained to the manifold during training, while its tangent readout provides a \(d\)-dimensional Euclidean vector to the downstream predictor.

For each categorical feature $s\in\mathcal{J}_{\mathrm{c}}$, HTNN assigns every value $c\in\mathcal{C}_s$ an HFV representation:
\begin{equation}
\small
\mathbf{z}^{\mathrm{c}}_{s,c}
\in\mathbb{H}_{\kappa}^{d},
\qquad c\in\mathcal{C}_s.
\label{eq:categorical_lorentz_representation}
\end{equation}
Given an observed value $x_s^{\mathrm{c}}$, categorical lookup followed by the tangent readout produces
\begin{equation}
\small
\mathbf{h}^{\mathrm{c}}_s
\bigl(x_s^{\mathrm{c}}\bigr)
=
\psi\!\left(
\mathbf{z}^{\mathrm{c}}_{s,x_s^{\mathrm{c}}}
\right)
\in\mathbb{R}^{d}.
\label{eq:categorical_tangent_representation}
\end{equation}

\subsection{Numerical HFV Representations}
\label{sec:numerical_representations}
Unlike a categorical feature, a numerical feature does not provide a fixed finite vocabulary for direct representation lookup.
HTNN therefore assigns each numerical feature $j\in\mathcal{J}_{\mathrm{n}}$ a feature-specific set of \(m_j\) learnable HFV representations:
\begin{equation}
\mathbf{Z}^{\mathrm{n}}_j
=
\left\{
\mathbf{z}^{\mathrm{n}}_{j,k}
\right\}_{k=1}^{m_j},
\qquad
\mathbf{z}^{\mathrm{n}}_{j,k}
\in\mathbb{H}_{\kappa}^{d}.
\label{eq:numerical_hfv_representations}
\end{equation}
Here, \(m_j\) is the HFV capacity of feature \(j\), namely, the number of feature-specific HFV representations assigned to that feature.
Each input value determines how their tangent readouts are combined into a single vector in \(\mathbb{R}^{d}\), matching the output space of the categorical pathway.

\paragraph{Feature-wise HFV capacity.}
Numerical features can differ substantially in empirical diversity and observation support.
HTNN therefore determines \(m_j\) once from the finite, non-missing training values of feature \(j\), without using labels.
Let \(\widehat{\mathcal{H}}_j\) denote their empirical entropy in bits, \(n_j\) the number of finite, non-missing training observations, and \(m_{\min}\) and \(m_{\max}\) the global capacity bounds.
For \(n_j>0\), we compute
\vspace{-5pt}
\begin{equation}
\small
\begin{aligned}
m_j^{\mathrm{ent}}
&=
\max\left\{
m_{\min},
\operatorname{round}\!\left(2^{\widehat{\mathcal{H}}_j}\right)
\right\},\\
m_j^{\mathrm{obs}}
&=
\max\left\{
m_{\min},
\left\lfloor\log_2 n_j\right\rfloor
\right\},\\
m_j
&=
\min\left\{
m_{\max},
m_j^{\mathrm{ent}},
m_j^{\mathrm{obs}}
\right\}.
\end{aligned}
\label{eq:adaptive_hfv_capacity}
\end{equation}
\vspace{-10pt}

The entropy-based term translates empirical diversity into a feature-specific capacity proposal, while the observation-based term limits it according to the available sample support.
Their minimum enforces both constraints within the global range \([m_{\min},m_{\max}]\).
This rule determines only the number of HFV representations assigned to each feature; it neither defines value bins nor associates individual representations with particular values or intervals.
The entropy estimator and edge-case handling are specified in Appendix~\ref{app:adaptive_capacity}.

\paragraph{Input-dependent composition.}
For an observed value \(x_j^{\mathrm{n}}\), HTNN generates one coefficient for each HFV representation:
\begin{equation}
\small
\boldsymbol{\alpha}_j(x_j^{\mathrm{n}})
=
\mathbf{a}_j x_j^{\mathrm{n}}+\mathbf{b}_j
\in\mathbb{R}^{m_j},
\label{eq:numerical_coefficients}
\end{equation}
where \(\mathbf{a}_j,\mathbf{b}_j\in\mathbb{R}^{m_j}\) are learnable Euclidean parameters.
The coefficients are neither normalized nor bounded, allowing numerical inputs to induce signed, scale-sensitive compositions.

Let
\(
\mathbf{v}^{\mathrm{n}}_{j,k}
=
\psi\!\left(\mathbf{z}^{\mathrm{n}}_{j,k}\right)
\in\mathbb{R}^{d}
\)
denote the tangent readout of the \(k\)-th HFV representation.
The resulting numerical tangent representation is
\vspace{-10pt}
\begin{equation}
\small
\mathbf{h}^{\mathrm{n}}_j(x_j^{\mathrm{n}})
=
\sum_{k=1}^{m_j}
\alpha_{j,k}(x_j^{\mathrm{n}})
\mathbf{v}^{\mathrm{n}}_{j,k}
\in\mathbb{R}^{d}.
\label{eq:numerical_composition}
\end{equation}
\vspace{-15pt}

\paragraph{Affine inference form.}
Because the coefficients are affine in \(x_j^{\mathrm{n}}\), the numerical composition reduces exactly to an affine map.
Defining
\(
\mathbf{u}_j=\sum_{k=1}^{m_j}
a_{j,k}\mathbf{v}^{\mathrm{n}}_{j,k}
\)
and
\(
\mathbf{c}_j=\sum_{k=1}^{m_j}
b_{j,k}\mathbf{v}^{\mathrm{n}}_{j,k}
\),
we obtain
\begin{equation}
\small
\mathbf{h}^{\mathrm{n}}_j(x_j^{\mathrm{n}})
=
\mathbf{u}_j x_j^{\mathrm{n}}+\mathbf{c}_j.
\label{eq:numerical_affine_form}
\end{equation}
This identity separates the affine numerical representation used for prediction from the HFV-based parameterization used during training.
Its optimization consequences and exact post-training conversion are analyzed in Section~\ref{sec:training_geometry_free_inference}.

\subsection{Prediction and Hybrid Optimization}
\label{sec:prediction_optimization}

\paragraph{Prediction.}
Using the categorical tangent representations in \eqref{eq:categorical_tangent_representation} and the numerical tangent representations in \eqref{eq:numerical_composition}, HTNN forms the input to its predictor as
\begin{equation}
\small
\mathbf{h}(\mathbf{x})
=
\operatorname{concat}\!\left(
\left\{
\mathbf{h}^{\mathrm{c}}_s(x_s^{\mathrm{c}})
\right\}_{s\in\mathcal{J}_{\mathrm{c}}},
\left\{
\mathbf{h}^{\mathrm{n}}_j(x_j^{\mathrm{n}})
\right\}_{j\in\mathcal{J}_{\mathrm{n}}}
\right)
\in
\mathbb{R}^{d\left(
|\mathcal{J}_{\mathrm{c}}|
+
|\mathcal{J}_{\mathrm{n}}|
\right)}.
\label{eq:heterogeneous_concatenation}
\end{equation}
The concatenated representation is passed to an MLP with a task-specific output layer, \(\widehat{y}=f_{\Theta_{\mathrm{p}}}(\mathbf{h}(\mathbf{x}))\), where \(\Theta_{\mathrm{p}}\) denotes the predictor parameters.
All parameters are trained jointly by minimizing the corresponding classification or regression loss \(\mathcal{L}\).

\paragraph{Hybrid optimization.}
HTNN separates its parameters according to their geometric domains:
\begin{equation}
\small
\begin{aligned}
\Theta_{\mathrm{H}}
&=
\left\{
\mathbf{z}^{\mathrm{c}}_{s,c}
\right\}_{s\in\mathcal{J}_{\mathrm{c}},\,c\in\mathcal{C}_s}
\cup
\left\{
\mathbf{z}^{\mathrm{n}}_{j,k}
\right\}_{j\in\mathcal{J}_{\mathrm{n}},\,1\leq k\leq m_j},
\\
\Theta_{\mathrm{E}}
&=
\left\{
\mathbf{a}_j,\mathbf{b}_j
\right\}_{j\in\mathcal{J}_{\mathrm{n}}}
\cup
\Theta_{\mathrm{p}}.
\end{aligned}
\label{eq:parameter_groups}
\end{equation}
As summarized in the center panel of Figure~\ref{fig:htnn_arch}, the HFV representations in \(\Theta_{\mathrm{H}}\) are optimized with RiemannianAdam, which maintains the Lorentz constraints, while the unconstrained Euclidean parameters \(\Theta_{\mathrm{E}}\) are optimized with AdamW.
The Lorentz scale \(\kappa\) remains fixed throughout training.

\subsection{Geometry-aware Training and Geometry-Free Inference Framework}
\label{sec:training_geometry_free_inference}

We now characterize how the HFV-based parameterization underlying \eqref{eq:numerical_affine_form} changes the effective affine update and how its geometric operations can be removed after training.

\paragraph{Induced affine dynamics.}
Let $\mathbf{B}_j
=
\begin{bmatrix}
\mathbf{u}_j & \mathbf{c}_j
\end{bmatrix}
\in\mathbb{R}^{d\times 2}$
collect the effective affine parameters of numerical feature \(j\).
Under continuous-time gradient flow, direct optimization of the affine map and HFV-based optimization in HTNN yield
\begin{equation}
\small
\dot{\mathbf{B}}_j
=
\begin{cases}
-\nabla_{\mathbf{B}_j}\mathcal{L},
& \text{direct affine optimization},\\[2mm]
-\mathcal{K}_j
\!\left[
\nabla_{\mathbf{B}_j}\mathcal{L}
\right],
& \text{HTNN}.
\end{cases}
\label{eq:induced_affine_dynamics}
\end{equation}
Here, \(\mathcal{K}_j\) is a state-dependent operator determined by the current HFV tangent readouts, their composition coefficients, and the local manifold geometry.
The two parameterizations can therefore realize the same affine map while following different optimization trajectories.

\paragraph{Curvature-dependent updates.}
The operator \(\mathcal{K}_j\) combines two update paths: the Euclidean coefficients change how the current HFV tangent readouts form \(\mathbf{u}_j\) and \(\mathbf{c}_j\), while the HFV representations change the readouts themselves.
Hyperbolic geometry enters through the latter path.

Consider an HFV representation \(\mathbf{z}\) with tangent readout \(\mathbf{v}=\psi(\mathbf{z})\neq\mathbf{0}\), and write \(\mathbf{v}=r\mathbf{q}\), where \(r=\lVert\mathbf{v}\rVert_2\) and \(\mathbf{q}=\mathbf{v}/r\).
For curvature \(-1/\kappa\), let \(\tau=\kappa^{-1/2}\).
In these normal coordinates, the transverse metric scale grows as \(\sinh(\tau r)/(\tau r)\), so the inverse metric increasingly attenuates transverse gradient components as \(r\) grows.
The inverse metric operator in these coordinates is
\begin{equation}
\small
P_{\mathrm{H}}(\mathbf{v})
=
\mathbf{q}\mathbf{q}^{\top}
+
\left(
\frac{\tau r}{\sinh(\tau r)}
\right)^2
\left(
I_d-\mathbf{q}\mathbf{q}^{\top}
\right),
\label{eq:hyperbolic_inverse_metric}
\end{equation}
with the continuous extension \(P_{\mathrm{H}}(\mathbf{0})=I_d\).
Accordingly, the HFV tangent readout evolves as
\(
\dot{\mathbf{v}}
=
-P_{\mathrm{H}}(\mathbf{v})
\nabla_{\mathbf{v}}\mathcal{L}
\).
The radial component is unchanged, whereas the transverse component is scaled by
\(
(\tau r/\sinh(\tau r))^2<1
\)
for \(r>0\).
Negative curvature therefore preserves radial responsiveness while increasingly attenuating transverse updates away from the origin.
We refer to this combined behavior as radial adaptability with transverse stabilization.

Categorical HFV representations are shaped directly by the same geometry-dependent update.
For numerical features, \(P_{\mathrm{H}}\) enters the HFV-update component of \(\mathcal{K}_j\), while the coefficient-update component remains Euclidean.
The exact derivations of \(P_{\mathrm H}\) and \(\mathcal K_j\) are provided in Appendix~\ref{app:optimization_analysis}.
Appendix~\ref{app:optimization_dynamics} further verifies the predicted radial and transverse responses and examines their manifestation in practical optimization trajectories.

\paragraph{Geometry-free inference.}
As shown in the right panel of Figure~\ref{fig:htnn_arch}, after training, each categorical tangent readout can be stored as an ordinary Euclidean lookup vector.
For each numerical feature, the HFV representations and composition coefficients can be replaced exactly by the effective parameters \(\mathbf{u}_j\) and \(\mathbf{c}_j\) in \eqref{eq:numerical_affine_form}.
Inference then requires only Euclidean lookup, affine transformation, and the downstream MLP; it contains no manifold operations or input-dependent HFV composition, and its cost is independent of \(m_j\).
This conversion leaves the model predictions unchanged.
HTNN thus realizes \emph{geometry-aware training and geometry-free inference}.

\section{Experiments}
\label{sec:experiments}

\subsection{Experimental Setup}
\label{sec:experimental_setup}

\paragraph{Benchmarks.}
We evaluate HTNN within the unified TALENT framework~\citep{talent}.
The full TALENT benchmark contains 300 datasets (200 classification and 100 regression tasks) and supports broad comparison across heterogeneous tabular problems.
TALENT-tiny is a compact 45-dataset core (30 classification and 15 regression tasks) curated under stricter quality and balance criteria for rapid, reproducible, and controlled in-depth evaluation.
Following these roles, we use the full benchmark for the main predictive comparison and TALENT-tiny for controlled analyses of optimization dynamics, architecture, feature-wise HFV capacity, and computational efficiency.

\paragraph{Compared methods.}
We compare HTNN with 35 baselines on classification and 33 baselines on regression, yielding fixed task-specific pools of 36 and 34 methods, respectively.
The pools span classical and tree-based methods, supervised neural architectures, retrieval-based methods, kernel-based feature learners, and tabular foundation models.
For the efficiency study, we evaluate the union of the ten highest-ranked methods in the full-benchmark classification and regression comparisons.
Full-benchmark baseline results are obtained either from TALENT's released result records or by executing the corresponding TALENT evaluation pipeline when no released record is available.
For efficiency evaluation, the selected methods are rerun under a common timing protocol using the same selected configurations or released pretrained checkpoints as in the predictive comparison, without repeating hyperparameter search.
The complete comparison pools, result provenance, and efficiency configurations are provided in Appendix~\ref{app:compared_methods}.

\paragraph{Evaluation protocol.}
We follow the official TALENT train, validation, and test splits, using accuracy (higher is better) for classification and root mean squared error (RMSE; lower is better) for regression.
For each model--dataset pair, we report the mean test score over 15 random runs, with all downstream model-selection decisions made exclusively on the validation split.
To aggregate performance across datasets, we rank methods within each dataset, with rank one denoting the best result, and use the mean rank as our primary summary statistic.
We accompany the mean rank with its across-dataset standard deviation and the full distribution of per-dataset ranks.
Pairwise win/tie/loss counts and multiplicity-adjusted statistical tests against all baselines are reported in Appendix~\ref{app:statistical_tests}, with the comparison protocol detailed in Appendix~\ref{app:pairwise_comparison_protocol}.
For the efficiency analysis, fitting and inference times are normalized per 1K training and test samples, respectively.
Each reported model-level value is the median across dataset-level means, where each dataset mean is computed over 15 runs.
Detailed timing definitions are provided in Appendix~\ref{app:efficiency_protocol}.

\paragraph{Implementation.}
We implement HTNN in PyTorch, using Geoopt for manifold operations and Riemannian optimization.
We fix the Lorentz scale to \(\kappa=1.0\) and use \(m_{\min}=2\) and \(m_{\max}=12\) for feature-wise HFV capacity.
Following TALENT's default neural-model training protocol, all HTNN runs use a batch size of 1,024, train for at most 200 epochs, and apply early stopping with a patience of 20 epochs.
For each dataset, we perform 100 hyperparameter-optimization trials using Optuna~\citep{akiba2019optuna}, searching over MLP depth, hidden width, and dropout rate, the representation dimension \(d\), the Euclidean and Riemannian learning rates, and weight decay.
Following Geoopt's recommendation for numerical stability in manifold computations, HTNN is trained and evaluated in double precision.\footnote{\url{https://geoopt.readthedocs.io/en/latest/extended/stereographic.html\#stability}}
The complete hyperparameter search space is provided in Appendix~\ref{app:htnn_search_space}.

\subsection{Main Results on TALENT}
\label{sec:main_results}

\begin{figure}[htbp]
	\small
    \centering
    \includegraphics[width=380pt]{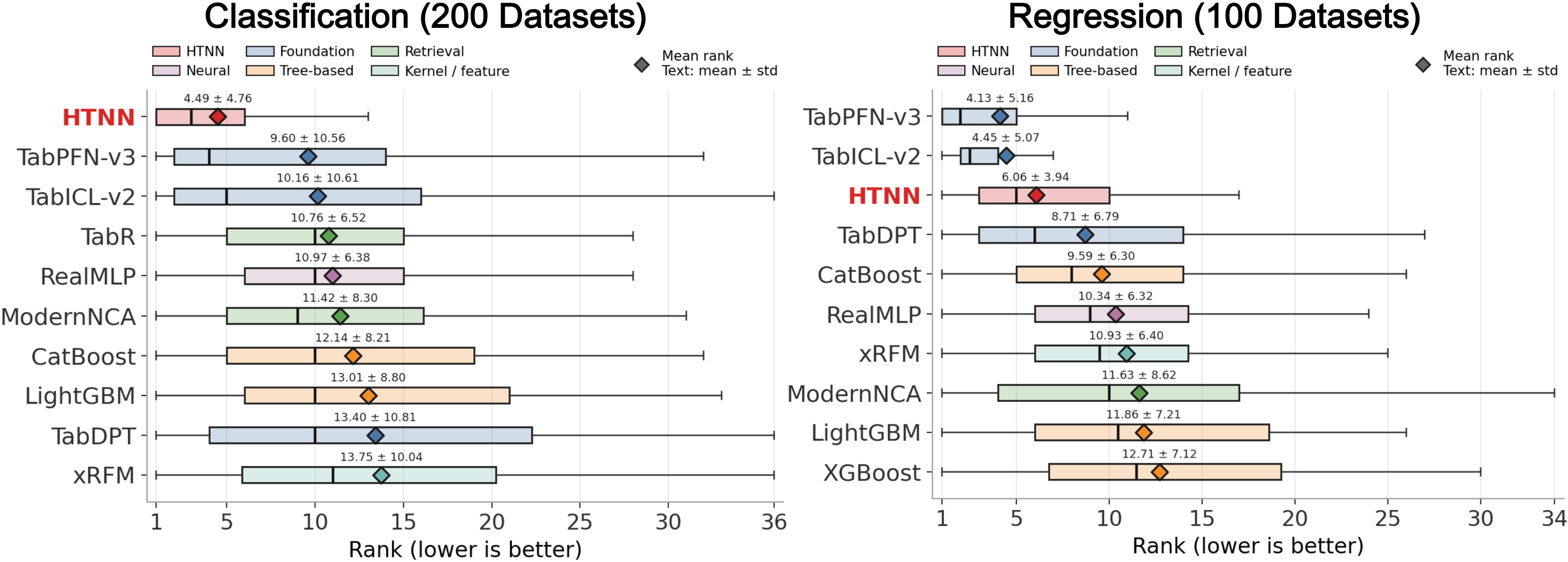}
    \vskip -10pt
    \caption{
    Per-dataset rank distributions on the full TALENT benchmark.
    Boxes summarize the rank distributions, and diamonds denote mean ranks. Annotations report the mean and across-dataset standard deviation.
    Lower ranks indicate better performance.
    Although only representative methods are displayed, ranks are computed using the complete task-specific comparison pools.
    }
    \label{fig:main_results}
\end{figure}

Figure~\ref{fig:main_results} summarizes the per-dataset rank distributions of HTNN and representative high-performing methods from the complete task-specific comparison pools.

\paragraph{Classification.}
HTNN ranks first among all 36 methods, with a mean rank of \(4.49\), more than five positions ahead of the second-ranked TabPFN-v3.
Its rank distribution remains concentrated near the top of the comparison pool across 200 datasets, showing that the aggregate advantage is broad rather than driven by a small subset of tasks.
HTNN also outperforms the strongest method from every other represented model family, including all three foundation models.
Pairwise tests confirm this result: after Holm correction, HTNN significantly outperforms each of the 35 classification baselines.
The complete W/T/L counts and Holm-adjusted \(p\)-values are reported in Appendix~\ref{app:statistical_tests}.

\paragraph{Regression.}
HTNN ranks third among the 34 methods, behind TabPFN-v3 and TabICL-v2, and is the highest-ranked non-foundation method.
Despite trailing these two pretrained models, it significantly outperforms each of the remaining 31 baselines, including TabDPT and the strongest alternatives from every non-foundation model family.
The complete W/T/L counts and Holm-adjusted \(p\)-values are reported in Appendix~\ref{app:statistical_tests}.

Taken together, HTNN delivers the strongest classification result and the strongest regression result among non-foundation methods, demonstrating its competitiveness across different tabular tasks.
Complete task-specific rankings and per-dataset scores are reported in Appendix~\ref{app:full_benchmark_results}.

\vskip -10pt
\vskip -10pt

\subsection{What Makes HTNN Work?}
\label{sec:architecture_ablation}

To isolate the contributions of HTNN's design choices, we compare it with eight controlled variants on TALENT-tiny, ranking all nine methods within a fixed variant pool.
Each variant is tuned independently using the same HPO budget, preprocessing pipeline, training protocol, and search space for shared hyperparameters.

\vskip -18pt
\begin{table}[H]
\small
\centering
\caption{
Controlled architecture ablation on TALENT-tiny.
Values are the mean and across-dataset standard deviation of ranks computed within the fixed nine-variant pool over 30 classification and 15 regression datasets.
Lower is better.
}
\label{tab:architecture_ablation}
\setlength{\tabcolsep}{5pt}
\renewcommand{\arraystretch}{1.08}
\begin{tabular}{@{}llllcc@{}}
\toprule
& \multicolumn{1}{l}{Categorical}
& \multicolumn{1}{l}{Numerical}
& \multicolumn{1}{l}{Coefficient}
& \multicolumn{2}{c}{Mean rank \(\downarrow\)} \\
\cmidrule(lr){5-6}
Variant
& representation
& module
& map
& Classification
& Regression \\
\midrule
\textbf{HTNN}
& Lorentz
& Lorentz composition
& Affine
& \(\mathbf{1.38 \pm 0.98}\)
& \(\mathbf{1.20 \pm 0.41}\) \\
\midrule
HTNN-Sphere
& Spherical
& Spherical composition
& Affine
& \(5.12 \pm 1.58\)
& \(4.67 \pm 1.23\) \\
HTNN-Euc
& Euclidean
& Euclidean composition
& Affine
& \(5.68 \pm 1.68\)
& \(5.53 \pm 1.06\) \\
\midrule
HTNN-CHNE
& Lorentz
& Euclidean composition
& Affine
& \(5.58 \pm 1.91\)
& \(5.60 \pm 2.16\) \\
HTNN-CENH
& Euclidean
& Lorentz composition
& Affine
& \(4.12 \pm 3.03\)
& \(2.40 \pm 1.18\) \\
\midrule
HTNN-CHNL
& Lorentz
& Direct affine
& ---
& \(4.38 \pm 1.88\)
& \(4.73 \pm 2.05\) \\
HTNN-CENL
& Euclidean
& Direct affine
& ---
& \(5.12 \pm 1.70\)
& \(4.73 \pm 2.28\) \\
\midrule
HTNN-Softmax
& Lorentz
& Lorentz composition
& Softmax
& \(6.65 \pm 1.60\)
& \(7.93 \pm 1.10\) \\
HTNN-Sigmoid
& Lorentz
& Lorentz composition
& Sigmoid
& \(6.97 \pm 1.83\)
& \(8.20 \pm 0.94\) \\
\bottomrule
\end{tabular}
\end{table}
\vspace{-25pt}

\paragraph{Controlled variants.}
HTNN-Sphere and HTNN-Euc replace the Lorentz geometry in both feature pathways with their matched spherical (curvature \(+1\)) and Euclidean counterparts, respectively, while HTNN uses Lorentz geometry with curvature \(-1\).
HTNN-CHNE retains hyperbolic categorical representations but uses a Euclidean numerical composition, whereas HTNN-CENH makes the complementary replacement.
HTNN-CHNL and HTNN-CENL replace the numerical composition module with the direct affine map in \eqref{eq:numerical_affine_form}, paired with hyperbolic and Euclidean categorical representations, respectively.
Finally, HTNN-Softmax and HTNN-Sigmoid retain the Lorentz numerical composition but apply the corresponding transformation to the coefficient vector in \eqref{eq:numerical_coefficients}.
Table~\ref{tab:architecture_ablation} summarizes the results; exact variant formulations and pairwise comparisons are provided in Appendix~\ref{app:architecture_ablation}.

\paragraph{Negative curvature.}
HTNN achieves the best mean rank on both tasks and significantly outperforms its matched spherical and Euclidean counterparts.
The spherical result shows that the gain is not a generic consequence of using a curved manifold, while the Euclidean result shows that the matched compositional parameterization alone is insufficient.
Together, these comparisons identify negative curvature as the effective geometric choice within HTNN.
A controlled optimization-dynamics analysis further tests the inverse-metric behavior predicted by \eqref{eq:hyperbolic_inverse_metric}: radial response remains unattenuated, whereas transverse response is increasingly attenuated with radius.
The measured local responses follow this prediction, with a maximum transverse-response error of \(7.2\times10^{-10}\).
The same directional effect persists in practical training trajectories: relative to HTNN-Euc, HTNN exhibits a lower transverse-update fraction on all 30 classification and all 15 regression datasets, with mean reductions of \(8.77\) and \(9.00\) percentage points, respectively.
Full results are provided in Appendix~\ref{app:optimization_dynamics}.

\paragraph{Heterogeneous feature pathways.}
HTNN-CHNE and HTNN-CENH each retain one hyperbolic feature pathway while replacing the other with its Euclidean counterpart.
Among the eight ablation variants, HTNN-CENH performs best on both tasks, indicating a particularly strong contribution from the numerical HFV pathway, especially on regression.
Nevertheless, neither single-hyperbolic-pathway variant matches the complete model, supporting complementary contributions from the categorical and numerical HFV pathways.

\paragraph{HFV-based training parameterization.}
HTNN-CHNL retains the same hyperbolic categorical pathway as HTNN but replaces the numerical HFV composition with a directly optimized affine map.
Although this variant has the same numerical inference function class as HTNN, it performs substantially worse on both tasks.
The gap therefore supports the benefit of learning the affine numerical map through HFV representations, consistent with the induced dynamics in \eqref{eq:induced_affine_dynamics}.

\paragraph{Coefficient constraints.}
HTNN-Softmax and HTNN-Sigmoid occupy the final two positions on both tasks.
Their degradation supports the use of unconstrained affine coefficients, which allow numerical inputs to produce signed and scale-sensitive compositions.

Across both tasks, HTNN significantly outperforms all eight variants after Holm correction.
The corresponding W/T/L counts and statistical results are reported in Appendix~\ref{app:architecture_ablation}.

\subsection{Feature-Wise Adaptive HFV Capacity}
\label{sec:capacity_analysis}

We examine whether using \eqref{eq:adaptive_hfv_capacity} for choosing the number \(m_j\) of HFV representations assigned to each numerical feature is more effective than using a single globally fixed representation capacity.
On TALENT-tiny, we compare adaptive HTNN, with \(m_j\in\{2,\ldots,12\}\), against hyperbolic fixed-\(m\) variants for \(m\in\{1,\ldots,12\}\) and the corresponding fixed-\(m\) HTNN-Euc variants.
All 25 variants are ranked within the same fixed comparison pool and tuned independently using the same HPO budget and training protocol.

\vskip -10pt
\begin{figure}[H]
    \small
    \centering
    \includegraphics[width=\textwidth]{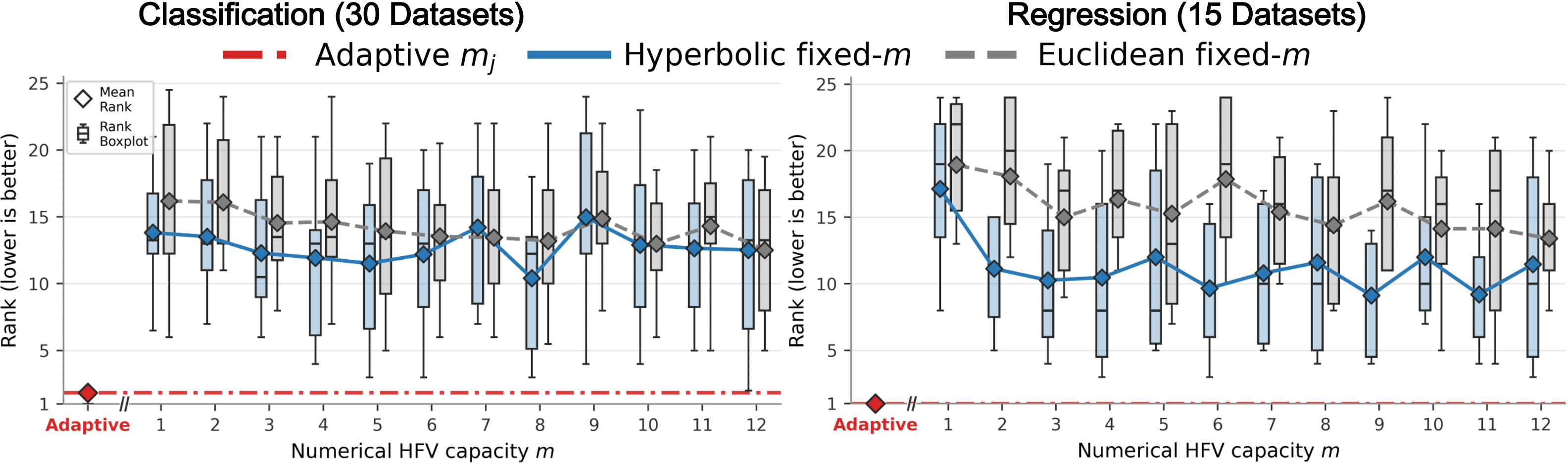}
    \vskip -10pt
    \caption{
    Feature-wise capacity analysis on TALENT-tiny.
    HTNN uses adaptive \(m_j\), whereas each fixed-capacity variant uses the same \(m\) for all numerical features.
    Boxplots summarize per-dataset ranks, and diamonds denote mean ranks.
    All ranks are computed within the fixed 25-variant pool.
    }
    \label{fig:adaptive_reference_capacity}
\end{figure}
\vspace{-20pt}

\paragraph{Feature-wise adaptive capacity.}

Adaptive HTNN outperforms every globally fixed alternative on both tasks, whereas fixed-\(m\) performance fluctuates rather than improving monotonically with \(m\). 
The capacities computed by \eqref{eq:adaptive_hfv_capacity} are highly heterogeneous across numerical features (see Appendix~\ref{app:capacity_results}): no single \(m_j\) accounts for more than approximately one third of the features, while a substantial fraction of features receive the minimum capacity. 
Thus, the adaptive rule neither approximates a single global \(m\) nor uniformly enlarges the representation layer. 
Since every adaptive \(m_j\) is bounded by the largest tested fixed \(m\), these results support the motivation behind \eqref{eq:adaptive_hfv_capacity}: matching each numerical feature's HFV capacity to its empirical information content yields better predictive performance than imposing a single global capacity.

\paragraph{Geometry under fixed capacity.}
At matched values of \(m\), the hyperbolic variants outperform their Euclidean counterparts for \(9/12\) settings on classification and \(12/12\) on regression.
This controlled pattern further supports the advantage of hyperbolic geometry observed in Section~\ref{sec:architecture_ablation}.

Complete ranking results and pairwise comparisons between adaptive HTNN and all 24 fixed-capacity variants are reported in Appendix~\ref{app:capacity_results}.

\subsection{Efficiency: Training Cost and Geometry-Free Inference}
\label{sec:efficiency}

\begin{figure}[H]
    \small
    \centering
    \includegraphics[width=\textwidth]{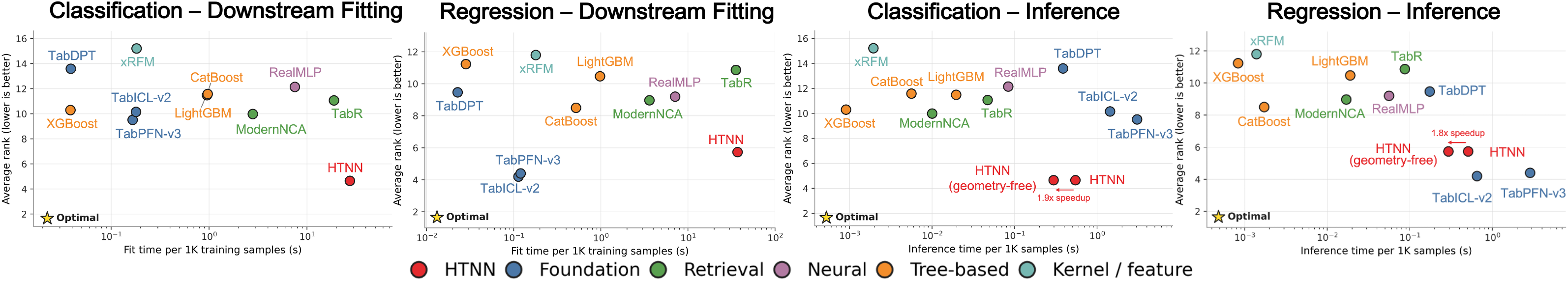}
    \vskip -10pt
    \caption{
    Performance versus downstream fitting and inference time on TALENT-tiny.
    Vertical axes show mean ranks over 30 classification or 15 regression datasets within the complete task-specific pools; horizontal axes show median time per 1K samples.
    The ``Optimal'' marker denotes the preferred lower-left region of low rank and runtime.
    HTNN (geometry-free) is the exact converted inference form of the same trained model.
    }
    \label{fig:efficiency}
\end{figure}

Figure~\ref{fig:efficiency} summarizes the trade-off between predictive performance and computational efficiency on TALENT-tiny.
For checkpoint-based foundation models, downstream fitting includes only their dataset-specific fitting stage and excludes pretraining. Per-dataset timing results are reported in Appendix~\ref{app:efficiency_results}.

\vskip -10pt
\vspace{-20pt}

\paragraph{Downstream fitting cost.}
The fitting panels make the computational trade-off explicit: HTNN achieves favorable ranks but incurs the largest median downstream fitting time among the displayed methods on both tasks.
This cost reflects \emph{geometry-aware training}, which combines Riemannian optimization of HFV representations with double-precision computation for numerical stability.

\paragraph{Geometry-free inference.}
We apply the exact conversion from Section~\ref{sec:training_geometry_free_inference} without retraining the model or changing its numerical precision.
The converted model matches the original HTNN's outputs up to a maximum absolute difference of \(5.47\times10^{-17}\) and retains the same predictive rank, while accelerating inference on all 45 TALENT-tiny datasets.
The median per-dataset speedup over the original HTNN is \(1.87\times\) for classification and \(1.84\times\) for regression.
Despite continuing to operate entirely in double precision, HTNN (geometry-free) reaches an inference time comparable to TabDPT on classification and remains clearly faster than TabICL-v2 and TabPFN-v3 on both tasks.
Thus, HTNN confines manifold-specific computation to \emph{geometry-aware training}, while exact conversion enables \emph{geometry-free inference} without changing predictions.

\section{Conclusion}
\label{sec:conclusion}

We introduced HTNN, a hyperbolic manifold constrained tabular neural network that learns categorical and numerical feature-value representations through \emph{geometry-aware training} and converts exactly to ordinary Euclidean operations for \emph{geometry-free inference}.
Across the full 300-dataset TALENT benchmark, HTNN ranks first on classification and third on regression.
Controlled studies support the roles of negative curvature, the complementary categorical and numerical pathways, and feature-wise HFV capacity, while the exact post-training conversion preserves predictions and accelerates inference.
Together, these results demonstrate the effectiveness of hyperbolic geometry for supervised tabular learning.

HTNN requires additional training cost due to Riemannian optimization and double-precision manifold computation, and the present study focuses on supervised learning with a conventional MLP predictor.
Future work will explore more efficient manifold optimization and extend hyperbolic feature-value representations to broader tabular architectures and learning settings.

\subsection*{AI Use Statement}

We used ChatGPT (Sol High) solely to assist with language polishing, review mathematical derivations for correctness, and provide suggestions for the design of ablation experiments.
All suggestions were critically reviewed by the authors, who made the final methodological decisions, conducted the experiments, and verified the results, and who take full responsibility for the content of this work.

\bibliography{iclr2027_conference}
\bibliographystyle{iclr2027_conference}

\appendix


\section{Optimization Analysis of the HFV-Based Training Parameterization}
\label{app:optimization_analysis}

We derive the normal-coordinate operator \(P_{\mathrm H}(\mathbf v)\) and the induced affine dynamics \(\mathcal K_j\) used in Section~\ref{sec:training_geometry_free_inference}.

\subsection{Normal-Coordinate Update}
\label{app:normal_coordinate_update}

Consider a tangent readout \(\mathbf v=\psi(\mathbf z)\neq\mathbf 0\) and write \(\mathbf v=r\mathbf q\), where \(r=\lVert\mathbf v\rVert_2\) is its geodesic distance from the origin and \(\mathbf q=\mathbf v/r\) is its radial direction.
For curvature \(-1/\kappa=-\tau^2\), where \(\tau=\kappa^{-1/2}\), the pullback metric in these normal coordinates is
\begin{equation}
g_{\mathrm H}(\mathbf v)
=
\mathbf q\mathbf q^\top
+
\left(
\frac{\sinh(\tau r)}{\tau r}
\right)^2
\left(
I_d-\mathbf q\mathbf q^\top
\right).
\end{equation}
The radial and transverse projectors are orthogonal.
Inverting their corresponding eigenvalues gives \(g_{\mathrm H}(\mathbf v)^{-1}=P_{\mathrm H}(\mathbf v)\), recovering \eqref{eq:hyperbolic_inverse_metric}.
At the origin, the metric factors converge to one, so
\(
g_{\mathrm H}(\mathbf 0)
=
P_{\mathrm H}(\mathbf 0)
=
I_d.
\)
Hence, for a loss expressed in the normal coordinate \(\mathbf v\), the corresponding Riemannian gradient flow is
\(
\dot{\mathbf v}
=
-
P_{\mathrm H}(\mathbf v)
\nabla_{\mathbf v}\mathcal L.
\)

\subsection{Induced Affine Dynamics}
\label{app:induced_affine_dynamics}

Fix a numerical feature \(j\).
For the \(k\)-th HFV representation, define its tangent readout and coefficient pair as
\(
\mathbf v^{\mathrm n}_{j,k}
=
\psi\!\left(\mathbf z^{\mathrm n}_{j,k}\right)
\in\mathbb R^d,
\boldsymbol{\beta}_{j,k}
=
\begin{bmatrix}
a_{j,k}\\
b_{j,k}
\end{bmatrix}
\in\mathbb R^2.
\)
The effective affine parameters in \eqref{eq:numerical_affine_form} can then be written as
\begin{equation}
\mathbf B_j
=
\begin{bmatrix}
\mathbf u_j & \mathbf c_j
\end{bmatrix}
=
\sum_{k=1}^{m_j}
\mathbf v^{\mathrm n}_{j,k}
\boldsymbol{\beta}_{j,k}^{\top}
\in\mathbb R^{d\times2}.
\end{equation}

Let \(\mathbf G_j=\nabla_{\mathbf B_j}\mathcal L \in\mathbb R^{d\times2}\), with the remaining model parameters held fixed.
The chain rule gives
\begin{equation}
\nabla_{\mathbf v^{\mathrm n}_{j,k}}\mathcal L
=
\mathbf G_j\boldsymbol{\beta}_{j,k},
\qquad
\nabla_{\boldsymbol{\beta}_{j,k}}\mathcal L
=
\mathbf G_j^\top\mathbf v^{\mathrm n}_{j,k}.
\end{equation}
Accordingly, the geometry-dependent tangent-readout flow and Euclidean coefficient flow are
\begin{equation}
\dot{\mathbf v}^{\mathrm n}_{j,k}
=
-
P_{\mathrm H}\!\left(\mathbf v^{\mathrm n}_{j,k}\right)
\mathbf G_j\boldsymbol{\beta}_{j,k},
\qquad
\dot{\boldsymbol{\beta}}_{j,k}
=
-
\mathbf G_j^\top\mathbf v^{\mathrm n}_{j,k}.
\end{equation}

Differentiating the HFV-based parameterization of \(\mathbf B_j\) and substituting these updates yields
\(\dot{\mathbf B}_j=-\mathcal K_j[\mathbf G_j]\), where, for any
\(\mathbf G\in\mathbb R^{d\times2}\),
\begin{equation}
\begin{aligned}
\mathcal K_j[\mathbf G]
&=
\left(
\sum_{k=1}^{m_j}
\mathbf v^{\mathrm n}_{j,k}
\bigl(\mathbf v^{\mathrm n}_{j,k}\bigr)^\top
\right)\mathbf G
\\
&\quad+
\sum_{k=1}^{m_j}
P_{\mathrm H}\!\left(\mathbf v^{\mathrm n}_{j,k}\right)
\mathbf G
\boldsymbol{\beta}_{j,k}
\boldsymbol{\beta}_{j,k}^{\top}.
\end{aligned}
\label{eq:app_induced_operator}
\end{equation}
The first term arises from coefficient updates and depends on the current HFV tangent readouts.
The second arises from HFV updates and additionally incorporates the geometry-dependent operator
\(P_{\mathrm H}(\mathbf v^{\mathrm n}_{j,k})\).
Consequently, \(\mathcal K_j\) evolves with both the HFV tangent readouts and their composition coefficients throughout training.

\noindent\textbf{Role of HFV capacity.}
When \(m_j=1\), this parameterization constrains \(\mathbf B_j\) to have rank at most one.
For \(m_j\geq2\), any \(\mathbf B_j=[\mathbf u_j,\mathbf c_j]\in\mathbb R^{d\times2}\) can be represented by choosing
\[
\mathbf v^{\mathrm n}_{j,1}=\mathbf u_j,
\qquad
\boldsymbol{\beta}_{j,1}=(1,0)^\top,
\qquad
\mathbf v^{\mathrm n}_{j,2}=\mathbf c_j,
\qquad
\boldsymbol{\beta}_{j,2}=(0,1)^\top,
\]
with any remaining contributions set to zero.
Additional HFV representations therefore provide alternative HFV-based parameterizations and modify \(\mathcal K_j\), without changing the affine inference form.

\noindent\textbf{Scope.}
The derivation uses unit-rate, unregularized continuous-time gradient flow to isolate the effects of the HFV-based training parameterization and manifold metric.
RiemannianAdam and AdamW additionally introduce adaptive rescaling, momentum, weight decay, and discrete updates, so the equations above do not describe their complete trajectories.
The analysis establishes that the two parameterizations induce different optimization dynamics.

\section{Feature-Wise HFV Capacity}
\label{app:adaptive_capacity}

For each numerical feature \(j\), HTNN converts its training entries to numerical values and removes missing and non-finite observations.
Let \(n_j\) denote the number of remaining valid observations.
The resulting HFV capacity \(m_j\) is computed once before model optimization, remains fixed throughout training, and uses neither labels nor validation or test data.
If \(n_j=0\), we set \(m_j=m_{\min}\).

\noindent\textbf{Entropy estimation.}
Let \(p_{j,b}\) denote the empirical probability of state or histogram bin \(b\) for numerical feature \(j\).
We estimate its entropy in bits as
\begin{equation}
\widehat{\mathcal H}_j
=
-
\sum_{b:p_{j,b}>0}
p_{j,b}\log_2 p_{j,b}.
\end{equation}
If the feature contains at most 30 distinct valid values, we treat them as discrete states and estimate \(p_{j,b}\) from their empirical frequencies.
Otherwise, we construct a histogram from the valid observations.
Its bin width is selected using the Freedman--Diaconis rule,
\begin{equation}
h
=
2\,\operatorname{IQR}\,n_j^{-1/3}.
\end{equation}
The histogram frequencies are then normalized to obtain \(p_{j,b}\).
Accordingly,
\(2^{\widehat{\mathcal H}_j}\)
represents the entropy-based effective number of empirical states rather than the raw number of distinct values.

\noindent\textbf{Numerical safeguards.}
A constant feature is handled by the discrete branch above and yields an entropy estimate of zero.
In the histogram branch, if the Freedman--Diaconis width is zero, as can occur when the interquartile range is zero, NumPy constructs a single-bin histogram, also yielding an entropy estimate of zero.
If the estimated Freedman--Diaconis histogram would contain more than 10,240 bins, we instead use 10,240 equal-width bins.
If histogram construction raises an exception, we fall back to 20 equal-width bins.
The resulting entropy estimate and \(n_j\) are substituted into \eqref{eq:adaptive_hfv_capacity} to determine \(m_j\).

\section{Experimental Details}
\label{app:experimental_details}

\subsection{Data Preprocessing}
\label{app:data_preprocessing}

All preprocessing statistics and encoders are fitted exclusively on the training split and then applied unchanged to the validation and test splits.
Missing numerical values are replaced with \(0.0\), after which numerical features are standardized using a \texttt{StandardScaler} fitted on the imputed training data.
Missing categorical values are represented by the dedicated token \texttt{\_\_missing\_\_}; the categorical encoder is fitted on the training split, and previously unseen categories are mapped to a dedicated unknown index.
The adaptive HFV capacity \(m_j\) is computed before these transformations using only the original finite, non-missing training values, without labels.
Classification label mappings and regression-target standardization statistics are likewise determined solely from the training split.

\subsection{Compared Methods and Result Provenance}
\label{app:compared_methods}

\paragraph{Full-benchmark comparison pools.}
Table~\ref{tab:full_comparison_pools} lists the complete task-specific comparison pools.
The classification pool contains 35 baselines and HTNN, while the regression pool contains 33 baselines and HTNN.

\begin{table}[!t]
\centering
\small
\setlength{\tabcolsep}{6pt}
\renewcommand{\arraystretch}{1.12}
\begin{tabularx}{\textwidth}{
    @{}
    >{\raggedright\arraybackslash}p{0.19\textwidth}
    >{\raggedright\arraybackslash}X
    @{}
}
\toprule
\multicolumn{2}{c}{\textbf{Classification (36 methods)}} \\
\midrule
\textbf{Family} & \textbf{Methods} \\
\midrule
Classical
&
Dummy, Logistic Regression, NCM, Naive Bayes, KNN, SVM
\\
\addlinespace[2pt]

Tree-based
&
XGBoost, CatBoost, Random Forest, LightGBM
\\
\addlinespace[2pt]

Supervised neural
&
MLP, ResNet, NODE, SwitchTab, TabNet, TabCaps, TANGOS,
DANets, FT-Transformer, AutoInt, DCNv2, SNN, TabTransformer,
PTaRL, GrowNet, MLP-PLR, RealMLP, ExcelFormer, TabM
\\
\addlinespace[2pt]

Retrieval-based
&
TabR, ModernNCA
\\
\addlinespace[2pt]

Kernel-based feature learning
&
xRFM
\\
\addlinespace[2pt]

Foundation
&
TabPFN-v3, TabICL-v2, TabDPT
\\
\addlinespace[2pt]

Proposed
&
HTNN
\\

\midrule
\multicolumn{2}{c}{\textbf{Regression (34 methods)}} \\
\midrule
\textbf{Family} & \textbf{Methods} \\
\midrule
Classical
&
Dummy, Linear Regression, KNN, SVM, DNNR
\\
\addlinespace[2pt]

Tree-based
&
XGBoost, CatBoost, Random Forest, LightGBM
\\
\addlinespace[2pt]

Supervised neural
&
MLP, ResNet, NODE, SwitchTab, TabNet, TANGOS, DANets,
FT-Transformer, AutoInt, DCNv2, SNN, TabTransformer, PTaRL,
GrowNet, MLP-PLR, RealMLP, ExcelFormer, TabM
\\
\addlinespace[2pt]

Retrieval-based
&
TabR, ModernNCA
\\
\addlinespace[2pt]

Kernel-based feature learning
&
xRFM
\\
\addlinespace[2pt]

Foundation
&
TabPFN-v3, TabICL-v2, TabDPT
\\
\addlinespace[2pt]

Proposed
&
HTNN
\\
\bottomrule
\end{tabularx}

\caption{Complete task-specific comparison pools used for the full-benchmark rankings.}
\label{tab:full_comparison_pools}
\end{table}

\paragraph{Result provenance.}
For baselines with complete prediction records released by TALENT, we use the corresponding scores directly.
For xRFM~\citep{xrfm} and TabM~\citep{tabm}, whose required full-benchmark scores were not available in the released result records, we execute their complete TALENT pipelines, including model-specific hyperparameter search, training, model selection, and test prediction.
For TabPFN-v3, TabICL-v2, and TabDPT~\citep{tabpfn3,tabiclv2,tabdpt}, we evaluate the released pretrained checkpoints through their corresponding TALENT interfaces.
Under this evaluation protocol, these foundation models require neither downstream hyperparameter search nor parameter training.

\paragraph{Efficiency configurations.}
For every trainable method, we reuse the configuration selected during its predictive evaluation.
For the foundation models, we reuse the same released checkpoint and inference configuration.
No additional hyperparameter search is performed or included in the efficiency measurements.

\subsection{Pairwise Comparison Protocol}
\label{app:pairwise_comparison_protocol}

For each task, we compare HTNN separately with every baseline over the corresponding datasets.
Win/tie/loss counts are determined from the model--dataset test scores averaged over 15 runs and rounded to four decimal places.
HTNN records a win when it achieves higher accuracy for classification or lower RMSE for regression, a loss under the opposite condition, and a tie when the rounded scores are equal.
If only one method has a valid score, the method with the valid score is counted as the winner; if both scores are missing, the comparison is counted as a tie.

For statistical testing, we use the same per-dataset ranks as in the full-benchmark summaries rather than raw score differences.
Raw scores do not provide a uniform scale across different datasets: in classification, the same five-percentage-point accuracy gain can represent substantially different error reductions at different performance levels (e.g., \(0.55\) to \(0.60\) versus \(0.95\) to \(1.00\)), while regression RMSE values and their absolute differences can vary by orders of magnitude with the target scale.
Per-dataset ranks remove these scale differences and align the statistical analysis with our primary mean-rank summary.
The ranks are computed from the unrounded mean scores using the complete task-specific comparison pool.
Rank one denotes the best result, tied scores receive average ranks, and missing results receive tied worst ranks.
We then apply a two-sided paired Wilcoxon signed-rank test to the per-dataset ranks of HTNN and each baseline, using the Pratt convention for zero differences.
To control the family-wise error rate, the resulting \(p\)-values are corrected with the Holm procedure independently over the 35 classification and 33 regression comparisons.
A comparison is considered significant when its Holm-adjusted \(p\)-value is below \(0.05\).

\subsection{Timing Protocol}
\label{app:efficiency_protocol}

All efficiency measurements are collected on a single machine equipped with an NVIDIA GeForce RTX 4080 GPU and an Intel Xeon Platinum 8481C CPU.
For GPU-based methods, the device is synchronized immediately before and after each timed region to account for asynchronous execution.
Tree-based methods are executed on the CPU, while the remaining methods are executed on the GPU.
Fitting and inference are timed separately over \(R=15\) runs, and all runs are retained.

Downstream fitting time is the elapsed time of the dataset-specific \texttt{fit} stage.
For trainable methods, this stage includes model optimization under the selected configuration.
For checkpoint-based foundation models, it corresponds to their dataset-specific \texttt{fit} call and excludes foundation-model pretraining.
Inference time is the elapsed time required to generate predictions for the complete test partition.

Let \(t^{\mathrm{fit}}_{i,d,r}\) and \(t^{\mathrm{inf}}_{i,d,r}\) denote the fitting and inference times of model \(i\) on dataset \(d\) in run \(r\).
Their dataset-level mean times, normalized to 1K rows, are
\begin{align}
\bar{t}^{\mathrm{fit}}_{i,d}
&=
\frac{10^3}{R N^{\mathrm{train}}_d}
\sum_{r=1}^{R} t^{\mathrm{fit}}_{i,d,r},
\\
\bar{t}^{\mathrm{inf}}_{i,d}
&=
\frac{10^3}{R N^{\mathrm{test}}_d}
\sum_{r=1}^{R} t^{\mathrm{inf}}_{i,d,r}.
\end{align}
The model-level values in Figure~\ref{fig:efficiency} are the medians of these dataset-level means over the corresponding classification or regression datasets.
The per-dataset tables additionally report the standard deviation across the 15 normalized run-level times.

\subsection{HTNN Hyperparameter Search Space}
\label{app:htnn_search_space}

\begin{table}[H]
\centering
\small
\renewcommand{\arraystretch}{1.15}

\scalebox{0.90}{%
\begin{tabular}{@{}llll@{}}
\toprule
\textbf{Hyperparameter}
& \textbf{Search space}
& \textbf{Hyperparameter}
& \textbf{Search space} \\
\midrule

MLP depth
& Integer in \([1,8]\)
& Hidden width per layer
& Integer in \([64,512]\) \\

Dropout rate per layer
& \(0\) or \(\operatorname{Uniform}(0,0.5)\)
& Representation dimension \(d\)
& Integer in \([4,32]\) \\

Euclidean learning rate
& \(\operatorname{LogUniform}(10^{-5},10^{-2})\)
& Riemannian learning rate
& \(\operatorname{LogUniform}(10^{-5},10^{-2})\) \\

Weight decay
& \(0\) or \(\operatorname{LogUniform}(10^{-6},10^{-3})\)
& &
\\

\bottomrule
\end{tabular}%
}

\caption{Hyperparameter search space for HTNN.}
\label{tab:htnn_search_space}
\end{table}

\section{Full-Benchmark Results}
\label{app:full_benchmark_results}

The full-benchmark result matrices are nearly complete: 17 of the 7,000 baseline--dataset entries are missing for classification, and 5 of the 3,300 entries are missing for regression.
HTNN has valid results on all 300 datasets.
We retain the fixed task-specific comparison pools without removing any method or dataset; missing entries are handled according to the protocol in Appendix~\ref{app:pairwise_comparison_protocol}.

\subsection{Pairwise Statistical Comparisons}
\label{app:statistical_tests}

Table~\ref{tab:full_benchmark_pairwise_tests} reports the complete pairwise comparisons between HTNN and every baseline in the task-specific pools.
On classification, all 35 comparisons favor HTNN and remain significant after Holm correction.
On regression, TabPFN-v3 and TabICL-v2 significantly outperform HTNN, whereas HTNN significantly outperforms each of the remaining 31 baselines.

\begin{table}[H]
    \centering
    \caption{
    Complete pairwise comparisons on the full TALENT benchmark.
    W/T/L counts are reported from the perspective of HTNN.
    Holm correction is applied independently over all 35 classification
    and 33 regression comparisons.
    }
    \label{tab:full_benchmark_pairwise_tests}

    \scriptsize
    \setlength{\tabcolsep}{2.5pt}
    \renewcommand{\arraystretch}{0.96}

    \begin{tabularx}{\textwidth}{
        @{}
        >{\raggedright\arraybackslash}X
        c
        c
        @{\hspace{1.0em}}
        >{\raggedright\arraybackslash}X
        c
        c
        @{}
    }
        \toprule
        \multicolumn{3}{c}{\textbf{Classification (200 datasets)}} &
        \multicolumn{3}{c}{\textbf{Regression (100 datasets)}} \\
        \cmidrule(lr){1-3}
        \cmidrule(lr){4-6}

        \textbf{Baseline} &
        \textbf{W/T/L} &
        \(\boldsymbol{p_{\mathrm{Holm}}}\) &
        \textbf{Baseline} &
        \textbf{W/T/L} &
        \(\boldsymbol{p_{\mathrm{Holm}}}\) \\
        \midrule

        TabPFN-v3 & 112/2/86 & \(1.40{\times}10^{-4}\) &
        TabPFN-v3 & 30/2/68 & \(3.56{\times}10^{-3}\) \\

        TabICL-v2 & 115/1/84 & \(2.44{\times}10^{-5}\) &
        TabICL-v2 & 28/2/70 & \(1.77{\times}10^{-3}\) \\

        TabR & 159/2/39 & \(1.21{\times}10^{-16}\) &
        TabDPT & 52/4/44 & \(1.49{\times}10^{-2}\) \\

        RealMLP & 158/2/40 & \(1.11{\times}10^{-18}\) &
        CatBoost & 59/2/39 & \(1.65{\times}10^{-3}\) \\

        ModernNCA & 153/1/46 & \(3.54{\times}10^{-15}\) &
        RealMLP & 62/5/33 & \(1.93{\times}10^{-5}\) \\

        CatBoost & 158/1/41 & \(1.38{\times}10^{-17}\) &
        xRFM & 76/2/22 & \(8.56{\times}10^{-9}\) \\

        LightGBM & 162/1/37 & \(1.07{\times}10^{-18}\) &
        ModernNCA & 65/2/33 & \(2.00{\times}10^{-5}\) \\

        TabDPT & 143/4/53 & \(2.56{\times}10^{-16}\) &
        LightGBM & 68/2/30 & \(7.75{\times}10^{-6}\) \\

        xRFM & 153/4/43 & \(5.08{\times}10^{-18}\) &
        XGBoost & 66/2/32 & \(2.75{\times}10^{-7}\) \\

        XGBoost & 161/3/36 & \(1.38{\times}10^{-19}\) &
        TabR & 71/3/26 & \(3.79{\times}10^{-7}\) \\

        FT-Transformer & 186/0/14 & \(7.05{\times}10^{-29}\) &
        FT-Transformer & 85/3/12 & \(8.40{\times}10^{-13}\) \\

        MLP-PLR & 179/1/20 & \(1.75{\times}10^{-27}\) &
        MLP-PLR & 82/3/15 & \(8.40{\times}10^{-13}\) \\

        ResNet & 189/2/9 & \(2.09{\times}10^{-30}\) &
        ExcelFormer & 88/2/10 & \(6.48{\times}10^{-15}\) \\

        DCNv2 & 191/1/8 & \(1.19{\times}10^{-30}\) &
        Random Forest & 76/2/22 & \(5.39{\times}10^{-11}\) \\

        MLP & 192/0/8 & \(4.10{\times}10^{-31}\) &
        PTaRL & 95/3/2 & \(3.32{\times}10^{-16}\) \\

        AutoInt & 195/1/4 & \(2.10{\times}10^{-32}\) &
        AutoInt & 95/2/3 & \(1.64{\times}10^{-16}\) \\

        SNN & 193/1/6 & \(1.12{\times}10^{-31}\) &
        NODE & 86/2/12 & \(2.06{\times}10^{-14}\) \\

        Random Forest & 179/2/19 & \(6.15{\times}10^{-28}\) &
        DCNv2 & 97/2/1 & \(1.41{\times}10^{-16}\) \\

        ExcelFormer & 187/0/13 & \(2.33{\times}10^{-30}\) &
        MLP & 98/2/0 & \(1.21{\times}10^{-16}\) \\

        TANGOS & 196/1/3 & \(2.10{\times}10^{-32}\) &
        ResNet & 96/2/2 & \(1.41{\times}10^{-16}\) \\

        TabCaps & 197/0/3 & \(8.49{\times}10^{-33}\) &
        TabM & 88/2/10 & \(3.11{\times}10^{-15}\) \\

        PTaRL & 195/1/4 & \(3.76{\times}10^{-32}\) &
        TANGOS & 97/2/1 & \(1.21{\times}10^{-16}\) \\

        DANets & 197/1/2 & \(4.99{\times}10^{-33}\) &
        SNN & 96/2/2 & \(1.83{\times}10^{-16}\) \\

        TabM & 183/1/16 & \(5.48{\times}10^{-29}\) &
        KNN & 92/2/6 & \(4.78{\times}10^{-16}\) \\

        NODE & 191/1/8 & \(3.35{\times}10^{-32}\) &
        TabNet & 98/2/0 & \(1.21{\times}10^{-16}\) \\

        Logistic Regression & 191/3/6 & \(1.12{\times}10^{-31}\) &
        DNNR & 99/1/0 & \(1.21{\times}10^{-16}\) \\

        TabTransformer & 194/0/6 & \(6.51{\times}10^{-32}\) &
        GrowNet & 98/2/0 & \(1.21{\times}10^{-16}\) \\

        KNN & 188/2/10 & \(1.40{\times}10^{-31}\) &
        SVM & 98/2/0 & \(1.21{\times}10^{-16}\) \\

        SVM & 191/1/8 & \(2.99{\times}10^{-32}\) &
        Linear Regression & 98/2/0 & \(1.21{\times}10^{-16}\) \\

        GrowNet & 199/0/1 & \(4.74{\times}10^{-33}\) &
        TabTransformer & 100/0/0 & \(1.21{\times}10^{-16}\) \\

        SwitchTab & 199/0/1 & \(4.57{\times}10^{-33}\) &
        Dummy & 100/0/0 & \(1.21{\times}10^{-16}\) \\

        TabNet & 199/0/1 & \(4.57{\times}10^{-33}\) &
        SwitchTab & 100/0/0 & \(1.21{\times}10^{-16}\) \\

        Naive Bayes & 200/0/0 & \(4.57{\times}10^{-33}\) &
        DANets & 100/0/0 & \(1.21{\times}10^{-16}\) \\

        Dummy & 199/1/0 & \(4.57{\times}10^{-33}\) &
        \multicolumn{3}{c}{} \\

        NCM & 199/0/1 & \(4.57{\times}10^{-33}\) &
        \multicolumn{3}{c}{} \\

        \bottomrule
    \end{tabularx}
\end{table}

\subsection{Complete Task-Specific Rankings}
\label{app:complete_rankings}

Table~\ref{tab:complete_task_rankings} reports the complete task-specific rankings used in the full-benchmark comparison.

\begin{table}[H]
    \centering
    \caption{
    Complete rankings on the full TALENT benchmark.
    Values are the mean and standard deviation of per-dataset ranks.
    Lower mean ranks indicate better overall performance.
    The proposed method is highlighted in bold.
    }
    \label{tab:complete_task_rankings}

    \small
    \setlength{\tabcolsep}{3pt}
    \renewcommand{\arraystretch}{1.04}

    \begin{tabularx}{\textwidth}{
        @{}
        r
        >{\raggedright\arraybackslash}X
        c
        @{\hspace{1.5em}}
        r
        >{\raggedright\arraybackslash}X
        c
        @{}
    }
        \toprule
        \multicolumn{3}{c}{\textbf{Classification (200 datasets)}} &
        \multicolumn{3}{c}{\textbf{Regression (100 datasets)}} \\
        \cmidrule(lr){1-3}
        \cmidrule(lr){4-6}

        \textbf{Rank} &
        \textbf{Method} &
        \textbf{Mean \(\boldsymbol{\pm}\) Std} &
        \textbf{Rank} &
        \textbf{Method} &
        \textbf{Mean \(\boldsymbol{\pm}\) Std} \\
        \midrule

         1 & \textbf{HTNN}       & \(\mathbf{4.4875 \pm 4.7568}\)
        & 1 & TabPFN-v3          & \(4.1300 \pm 5.1594\) \\

         2 & TabPFN-v3           & \(9.5950 \pm 10.5644\)
        & 2 & TabICL-v2          & \(4.4500 \pm 5.0700\) \\

         3 & TabICL-v2           & \(10.1575 \pm 10.6116\)
        & 3 & \textbf{HTNN}      & \(\mathbf{6.0600 \pm 3.9359}\) \\

         4 & TabR                & \(10.7550 \pm 6.5242\)
        & 4 & TabDPT             & \(8.7100 \pm 6.7917\) \\

         5 & RealMLP             & \(10.9700 \pm 6.3779\)
        & 5 & CatBoost           & \(9.5900 \pm 6.3047\) \\

         6 & ModernNCA           & \(11.4175 \pm 8.3045\)
        & 6 & RealMLP            & \(10.3400 \pm 6.3229\) \\

         7 & CatBoost            & \(12.1450 \pm 8.2052\)
        & 7 & xRFM               & \(10.9300 \pm 6.4012\) \\

         8 & LightGBM            & \(13.0125 \pm 8.7968\)
        & 8 & ModernNCA          & \(11.6300 \pm 8.6224\) \\

         9 & TabDPT              & \(13.3975 \pm 10.8054\)
        & 9 & LightGBM           & \(11.8550 \pm 7.2105\) \\

        10 & xRFM                & \(13.7525 \pm 10.0369\)
        & 10 & XGBoost           & \(12.7150 \pm 7.1238\) \\

        11 & XGBoost             & \(13.8250 \pm 8.7183\)
        & 11 & TabR              & \(12.8800 \pm 8.1829\) \\

        12 & FT-Transformer      & \(14.7925 \pm 7.3149\)
        & 12 & FT-Transformer    & \(14.3650 \pm 7.3649\) \\

        13 & MLP-PLR             & \(15.0900 \pm 6.8650\)
        & 13 & MLP-PLR           & \(14.9300 \pm 7.9035\) \\

        14 & ResNet              & \(15.9900 \pm 7.2203\)
        & 14 & ExcelFormer       & \(15.3550 \pm 6.5595\) \\

        15 & DCNv2               & \(16.5650 \pm 7.2228\)
        & 15 & Random Forest     & \(15.4950 \pm 8.1496\) \\

        16 & MLP                 & \(16.7725 \pm 6.7343\)
        & 16 & PTaRL             & \(15.8250 \pm 5.5470\) \\

        17 & AutoInt             & \(17.0850 \pm 6.0929\)
        & 17 & AutoInt           & \(16.9900 \pm 5.9144\) \\

        18 & SNN                 & \(17.6700 \pm 6.2441\)
        & 18 & NODE              & \(17.1450 \pm 8.1984\) \\

        19 & Random Forest       & \(17.8925 \pm 9.2485\)
        & 19 & DCNv2             & \(17.8050 \pm 7.8852\) \\

        20 & ExcelFormer         & \(18.3575 \pm 7.4587\)
        & 20 & MLP               & \(17.9200 \pm 5.6469\) \\

        21 & TANGOS              & \(18.7550 \pm 6.6708\)
        & 21 & ResNet            & \(18.1850 \pm 6.5016\) \\

        22 & TabCaps             & \(18.8600 \pm 6.4329\)
        & 22 & TabM              & \(18.8700 \pm 7.3741\) \\

        23 & PTaRL               & \(19.5050 \pm 6.8472\)
        & 23 & TANGOS            & \(18.9600 \pm 5.9952\) \\

        24 & DANets              & \(19.8725 \pm 6.7041\)
        & 24 & SNN               & \(19.7550 \pm 6.0784\) \\

        25 & TabM                & \(21.5500 \pm 10.7237\)
        & 25 & KNN               & \(21.5450 \pm 7.1852\) \\

        26 & NODE                & \(21.9050 \pm 9.7890\)
        & 26 & TabNet            & \(24.2500 \pm 4.6103\) \\

        27 & Logistic Regression & \(22.4025 \pm 9.7989\)
        & 27 & DNNR              & \(24.9800 \pm 7.0061\) \\

        28 & TabTransformer      & \(22.5725 \pm 8.6433\)
        & 28 & GrowNet           & \(25.6950 \pm 4.4670\) \\

        29 & KNN                 & \(23.2075 \pm 9.7304\)
        & 29 & SVM               & \(25.8450 \pm 5.1868\) \\

        30 & SVM                 & \(24.4775 \pm 9.1442\)
        & 30 & Linear Regression & \(26.0850 \pm 5.0071\) \\

        31 & GrowNet             & \(25.9450 \pm 7.1921\)
        & 31 & TabTransformer    & \(29.9100 \pm 5.8652\) \\

        32 & SwitchTab           & \(27.1575 \pm 5.6443\)
        & 32 & Dummy             & \(30.3750 \pm 4.9074\) \\

        33 & TabNet              & \(28.0375 \pm 5.2934\)
        & 33 & SwitchTab         & \(30.4550 \pm 4.3351\) \\

        34 & Naive Bayes         & \(31.9400 \pm 5.7296\)
        & 34 & DANets            & \(30.9700 \pm 5.8109\) \\

        35 & Dummy               & \(32.8775 \pm 7.1158\)
        & \multicolumn{3}{c}{} \\

        36 & NCM                 & \(33.2050 \pm 4.2374\)
        & \multicolumn{3}{c}{} \\

        \bottomrule
    \end{tabularx}
\end{table}

\subsection{Per-Dataset HTNN Results}
\label{app:per_dataset_results}

Table~\ref{tab:htnn_per_dataset_results} reports HTNN's per-dataset test results on the full TALENT benchmark.
For compact presentation, we use the concise dataset display names adopted in Table~9 of \citet{tabpfnv2}; this changes only their presentation.

\input{tables/htnn_per_dataset_results.tex}

\section{Architecture Ablation Details}
\label{app:architecture_ablation}

\paragraph{Exact variant formulations.}
For a numerical representation geometry
\(G\in\{\mathbb H,\mathbb S,E\}\), define
\begin{equation}
\mathbf v^{G}_{j,k}
=
\begin{cases}
\psi_{\mathbb H}\!\left(\mathbf z^{\mathbb H}_{j,k}\right),
& G=\mathbb H,\\
\psi_{\mathbb S}\!\left(\mathbf z^{\mathbb S}_{j,k}\right),
& G=\mathbb S,\\
\mathbf v^{E}_{j,k},
& G=E,
\end{cases}
\end{equation}
where
\(
\mathbf z^{\mathbb H}_{j,k}\in\mathbb H^{d}_{1},
\mathbf z^{\mathbb S}_{j,k}\in\mathbb S^{d},
\mathbf v^{E}_{j,k}\in\mathbb R^{d}
\)
are learnable Lorentz, spherical, and Euclidean representations, respectively.
Using the affine coefficients from \eqref{eq:numerical_coefficients}, the corresponding composition-based numerical representation is
\begin{equation}
\mathbf h^{\mathrm n,G}_{j}(x_j^{\mathrm n})
=
\sum_{k=1}^{m_j}
\left(
a_{j,k}x_j^{\mathrm n}+b_{j,k}
\right)
\mathbf v^{G}_{j,k}.
\label{eq:app_geometry_reference_variants}
\end{equation}
All composition-based variants use the same adaptive \(m_j\) rule with
\(m_{\min}=2\) and \(m_{\max}=12\).

HTNN, HTNN-Sphere, and HTNN-Euc use the matched module pairs
\((\mathbb H,\mathbb H)\), \((\mathbb S,\mathbb S)\), and \((E,E)\),
respectively, where each pair specifies the categorical and numerical representation geometries.
HTNN-CHNE uses \((\mathbb H,E)\), whereas HTNN-CENH uses \((E,\mathbb H)\).
HTNN-CHNL and HTNN-CENL replace the numerical composition module with the direct affine map in \eqref{eq:numerical_affine_form}, paired with Lorentz and Euclidean categorical representations, respectively.

HTNN-Softmax and HTNN-Sigmoid retain Lorentz representations in both pathways but replace the unconstrained coefficients with
\begin{align}
\mathbf h^{\mathrm n,\mathrm{softmax}}_{j}(x_j^{\mathrm n})
&=
\sum_{k=1}^{m_j}
\left[
\operatorname{softmax}
\left(
\mathbf a_j x_j^{\mathrm n}+\mathbf b_j
\right)
\right]_{k}
\mathbf v^{\mathbb H}_{j,k},
\\
\mathbf h^{\mathrm n,\mathrm{sigmoid}}_{j}(x_j^{\mathrm n})
&=
\sum_{k=1}^{m_j}
\left[
\operatorname{sigmoid}
\left(
\mathbf a_j x_j^{\mathrm n}+\mathbf b_j
\right)
\right]_{k}
\mathbf v^{\mathbb H}_{j,k}.
\end{align}
Lorentz and spherical representations are optimized with RiemannianAdam, while Euclidean representations, composition coefficients, and predictor parameters are optimized with AdamW.

\paragraph{Pairwise comparisons.}
We follow the score, ranking, and Wilcoxon definitions in Appendix~\ref{app:pairwise_comparison_protocol}.
Ranks are computed within the fixed nine-variant pool, and Holm correction is applied independently over the eight HTNN--variant comparisons for each task.
All variants produce valid results on every TALENT-tiny dataset.

\begin{table}[H]
\centering
\caption{
Pairwise comparisons between HTNN and its controlled variants.
W/T/L reports the number of datasets on which HTNN wins, ties, or
loses against the listed variant.
Holm \(p\) denotes the multiplicity-adjusted \(p\)-value.
All comparisons remain significant at the \(0.05\) level after
correction.
}
\label{tab:architecture_pairwise_results}
\small
\setlength{\tabcolsep}{4.5pt}
\renewcommand{\arraystretch}{1.08}
\begin{tabular}{@{}lccc@{\hspace{1em}}ccc@{}}
\toprule
& \multicolumn{3}{c}{Classification}
& \multicolumn{3}{c}{Regression} \\
\cmidrule(lr){2-4}
\cmidrule(lr){5-7}
Variant
& W/T/L & Raw \(p\) & Holm \(p\)
& W/T/L & Raw \(p\) & Holm \(p\) \\
\midrule
HTNN-Sphere
& 28/2/0
& \(1.859{\times}10^{-6}\)
& \(1.487{\times}10^{-5}\)
& 15/0/0
& \(5.897{\times}10^{-4}\)
& \(4.371{\times}10^{-3}\) \\

HTNN-Euc
& 28/2/0
& \(1.917{\times}10^{-6}\)
& \(1.487{\times}10^{-5}\)
& 15/0/0
& \(5.882{\times}10^{-4}\)
& \(4.371{\times}10^{-3}\) \\

HTNN-CHNE
& 28/2/0
& \(1.928{\times}10^{-6}\)
& \(1.487{\times}10^{-5}\)
& 15/0/0
& \(6.354{\times}10^{-4}\)
& \(4.371{\times}10^{-3}\) \\

HTNN-CENH
& 23/4/3
& \(1.976{\times}10^{-5}\)
& \(1.976{\times}10^{-5}\)
& 12/0/3
& \(1.052{\times}10^{-2}\)
& \(1.052{\times}10^{-2}\) \\

HTNN-CHNL
& 27/2/1
& \(3.262{\times}10^{-6}\)
& \(1.487{\times}10^{-5}\)
& 15/0/0
& \(6.273{\times}10^{-4}\)
& \(4.371{\times}10^{-3}\) \\

HTNN-CENL
& 28/2/0
& \(1.885{\times}10^{-6}\)
& \(1.487{\times}10^{-5}\)
& 15/0/0
& \(6.241{\times}10^{-4}\)
& \(4.371{\times}10^{-3}\) \\

HTNN-Softmax
& 28/2/0
& \(1.875{\times}10^{-6}\)
& \(1.487{\times}10^{-5}\)
& 15/0/0
& \(5.464{\times}10^{-4}\)
& \(4.371{\times}10^{-3}\) \\

HTNN-Sigmoid
& 28/2/0
& \(1.920{\times}10^{-6}\)
& \(1.487{\times}10^{-5}\)
& 15/0/0
& \(5.655{\times}10^{-4}\)
& \(4.371{\times}10^{-3}\) \\
\bottomrule
\end{tabular}
\end{table}

\section{Controlled Optimization-Dynamics Analysis}
\label{app:optimization_dynamics}

\paragraph{Controlled setting.}
We examine whether the update behavior predicted by the hyperbolic metric appears during practical optimization.
We compare HTNN with the matched HTNN-Euc counterpart defined in Appendix~\ref{app:architecture_ablation} on all 45 TALENT-tiny datasets, comprising 30 classification and 15 regression tasks.
For each dataset and seed, the two models use matched initial tangent-coordinate representations, numerical coefficients, predictor parameters, and batch orders, and are trained for the same prespecified dataset-specific horizon.
Each model--dataset configuration is run with three seeds.
To make update magnitudes directly comparable, the learning rates of HTNN's manifold representations, its Euclidean parameters, and all HTNN-Euc parameters are set to the same value.
All other settings follow the selected dataset-specific HTNN configuration.
Because manifold and Euclidean representations are optimized with RiemannianAdam and AdamW, respectively, the trajectory comparison evaluates the realized end-to-end updates of the two training procedures rather than geometry under an identical optimizer, while the JVP analysis separately tests the local metric response.

\paragraph{Measurements.}
For a tangent-coordinate representation \(\mathbf v=r\mathbf q\), \eqref{eq:hyperbolic_inverse_metric} predicts an unattenuated radial response and a transverse response of
\(
R_{\perp}=(r/\sinh r)^2
\)
under HTNN's curvature of \(-1\).
We measure the realized local response using a Jacobian--vector product through the implemented manifold operations.
Dataset-level predicted and measured responses use the same transverse-gradient-energy weighting rather than evaluating the response at an average radius.

For the practical trajectories, we record each optimizer update
\(
\Delta\mathbf v_{t,i}
=
\mathbf v^{+}_{t,i}-\mathbf v^{-}_{t,i}
\)
for each active representation with a non-negligible pre-update norm, and decompose it relative to the pre-update radial direction
\(
\mathbf q_{t,i}
=
\mathbf v^{-}_{t,i}/\lVert\mathbf v^{-}_{t,i}\rVert_2
\).
Its transverse component is
\[
\Delta\mathbf v_{\perp,t,i}
=
\left(
I_d-\mathbf q_{t,i}\mathbf q_{t,i}^{\top}
\right)
\Delta\mathbf v_{t,i}.
\]
The transverse-update fraction is
\begin{equation}
F_{\perp}
=
\frac{
\sum_{t,i}
\left\lVert
\Delta\mathbf v_{\perp,t,i}
\right\rVert_2^2
}{
\sum_{t,i}
\left\lVert
\Delta\mathbf v_{t,i}
\right\rVert_2^2
},
\label{eq:transverse_update_fraction}
\end{equation}
where \(i\) indexes feature-value representations with active gradients.
Task-level means are computed from dataset-level results averaged over three seeds, and the 95\% confidence intervals are obtained from 10,000 bootstrap resamples of datasets.

\begin{figure}[H]
    \centering
    \includegraphics[width=\textwidth]{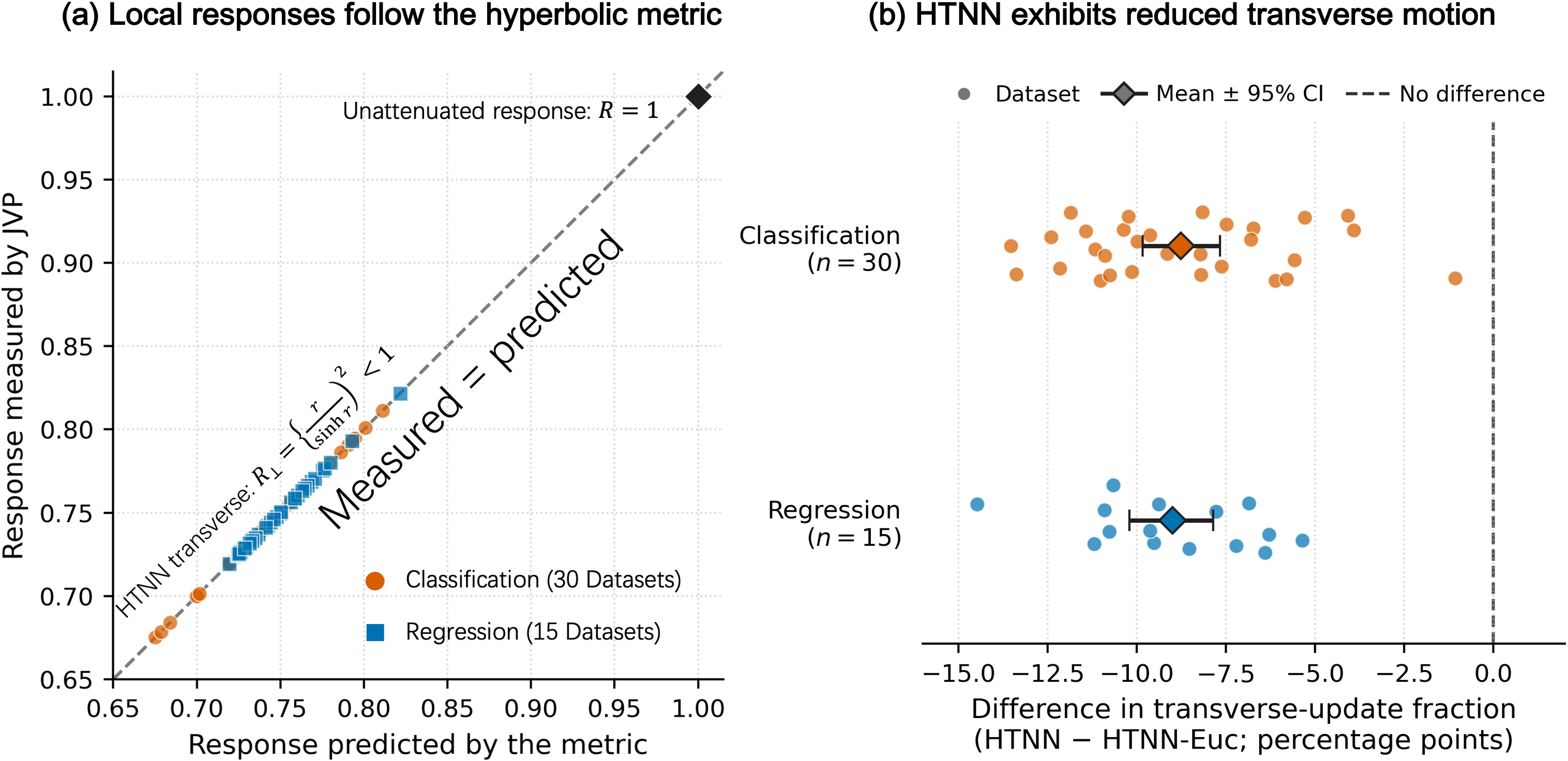}
    \caption{
    Controlled optimization-dynamics analysis on the 45 TALENT-tiny datasets.
    Each colored marker reports a dataset-level result averaged over three matched seeds.
    \textbf{(a)} JVP-measured local transverse responses closely follow the predictions of the hyperbolic metric.
    The black diamond marks the unattenuated response \(R=1\), predicted for HTNN's radial direction and for both directions under Euclidean geometry.
    \textbf{(b)} Difference in transverse-update fraction between HTNN and HTNN-Euc.
    Negative values indicate less transverse motion under HTNN.
    Diamonds and horizontal bars denote the task-level mean and its 95\% bootstrap confidence interval.
    }
    \label{fig:optimization_dynamics}
\end{figure}

\paragraph{Results.}
Figure~\ref{fig:optimization_dynamics}(a) shows that the measured transverse responses follow the metric predictions across all datasets, with a maximum absolute error of \(7.2\times10^{-10}\).
HTNN's radial response and both HTNN-Euc responses agree with the unattenuated value \(R=1\), whereas HTNN's transverse responses are consistently below one.
Figure~\ref{fig:optimization_dynamics}(b) shows that this directional effect persists in practical optimizer trajectories: HTNN has a lower transverse-update fraction than HTNN-Euc on all 30 classification and all 15 regression datasets.
The mean difference is \(-8.77\) percentage points for classification, with a 95\% confidence interval of \([-9.83,-7.67]\), and \(-9.00\) percentage points for regression, with an interval of \([-10.21,-7.85]\).
These observations connect the inverse-metric behavior presented in Section~\ref{sec:training_geometry_free_inference} and derived in Appendix~\ref{app:normal_coordinate_update} to the trajectories observed during training.
Together with HTNN's predictive advantage over its matched Euclidean counterpart, they support radial adaptability with transverse stabilization as the optimization behavior induced by HTNN's hyperbolic representation geometry.

\section{Capacity-Allocation Results}
\label{app:capacity_results}

Here, HFV capacity denotes the number of feature-specific HFV representations assigned to each numerical feature: adaptive HTNN computes a feature-specific \(m_j\), whereas each fixed-capacity variant uses the same \(m\) for all numerical features.

\begin{figure}[H]
    \centering
    \includegraphics[width=\textwidth]{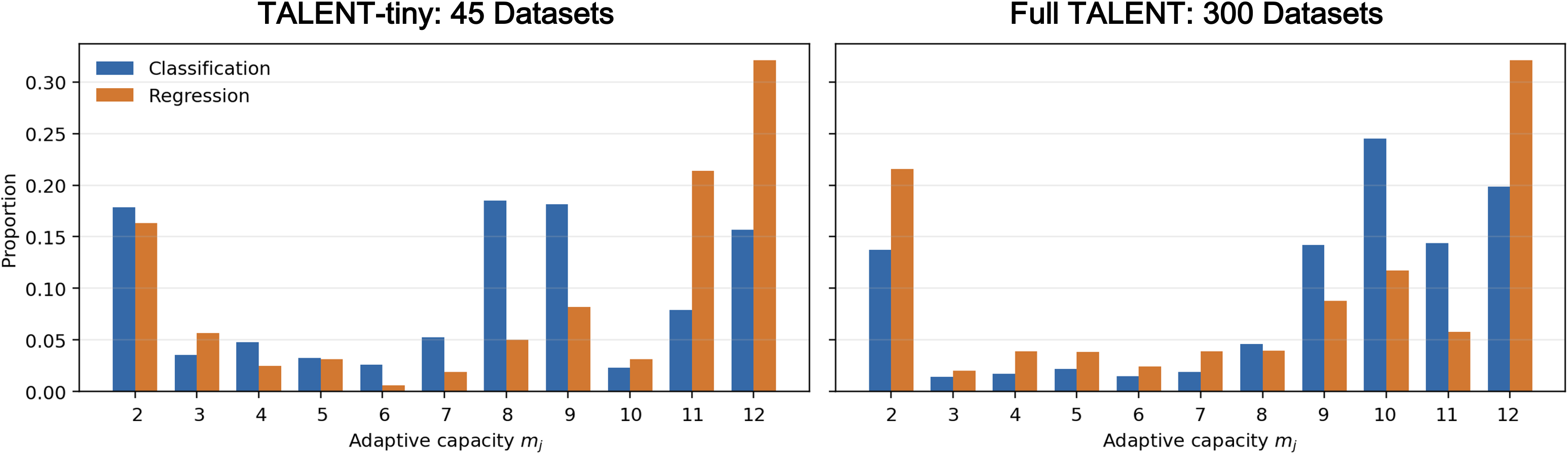}
    \caption{
    Distribution of the feature-wise HFV capacities \(m_j\) computed by \eqref{eq:adaptive_hfv_capacity}.
    The left panel covers the 45 TALENT-tiny datasets, and the right panel covers all 300 TALENT datasets.
    For each task, bar height denotes the proportion of numerical features assigned the corresponding capacity.
    }
    \label{fig:adaptive_capacity_distribution}
\end{figure}

Figure~\ref{fig:adaptive_capacity_distribution} shows the capacities computed by \eqref{eq:adaptive_hfv_capacity}.
The resulting allocations are highly heterogeneous across benchmark scales and prediction tasks: no single \(m_j\) accounts for more than approximately one third of the numerical features, and even the two most frequent capacities together cover fewer than \(60\%\), while a substantial fraction of features receive the minimum capacity \(m_j=2\).
Thus, the adaptive rule neither collapses to a nearly fixed capacity nor uniformly enlarges the HFV representation layer.
Together with adaptive HTNN's consistent advantage over all fixed-capacity variants, these distributions support feature-wise capacity allocation, rather than greater uniform capacity, as the explanation for the observed improvement.

Tables~\ref{tab:capacity_classification_results} and~\ref{tab:capacity_regression_results} report the complete fixed-capacity comparisons on TALENT-tiny.
For each fixed-capacity variant, W/T/L denotes the number of datasets on which adaptive HTNN wins, ties, or loses against that variant.
Ranks are computed within the fixed 25-variant pool, and Holm correction is applied independently over the 24 comparisons for each task.

\begin{table}[H]
\centering
\caption{
Complete capacity-allocation results on the 30 TALENT-tiny classification datasets.
Rank entries report the mean and across-dataset standard deviation within the fixed 25-variant pool. Lower is better.
Underlining marks the best fixed capacity within each geometry.
}
\label{tab:capacity_classification_results}
\scriptsize
\setlength{\tabcolsep}{3.0pt}
\renewcommand{\arraystretch}{1.08}
\begin{tabular*}{\textwidth}{
@{\extracolsep{\fill}}
c
cccc
cccc
@{}
}
\toprule
&
\multicolumn{4}{c}{Hyperbolic HTNN}
&
\multicolumn{4}{c}{HTNN-Euc}
\\
\cmidrule(lr){2-5}
\cmidrule(lr){6-9}
Capacity
& Mean rank
& W/T/L
& Raw \(p\)
& Holm \(p\)
& Mean rank
& W/T/L
& Raw \(p\)
& Holm \(p\)
\\
\midrule
\textbf{Adaptive \(m_j\)}
& \(\mathbf{1.83 \pm 3.04}\)
& ---
& ---
& ---
& \multicolumn{4}{c}{---}
\\
\midrule

\(m=1\)
& \(13.82 \pm 5.57\)
& 27/2/1
& \(2.667{\times}10^{-6}\)
& \(4.693{\times}10^{-5}\)
& \(16.18 \pm 6.51\)
& 28/2/0
& \(1.983{\times}10^{-6}\)
& \(4.693{\times}10^{-5}\)
\\

\(m=2\)
& \(13.52 \pm 5.95\)
& 28/2/0
& \(1.972{\times}10^{-6}\)
& \(4.693{\times}10^{-5}\)
& \(16.08 \pm 5.86\)
& 28/2/0
& \(1.995{\times}10^{-6}\)
& \(4.693{\times}10^{-5}\)
\\

\(m=3\)
& \(12.28 \pm 5.95\)
& 28/2/0
& \(1.956{\times}10^{-6}\)
& \(4.693{\times}10^{-5}\)
& \(14.53 \pm 5.34\)
& 28/2/0
& \(1.984{\times}10^{-6}\)
& \(4.693{\times}10^{-5}\)
\\

\(m=4\)
& \(11.93 \pm 5.95\)
& 28/2/0
& \(1.978{\times}10^{-6}\)
& \(4.693{\times}10^{-5}\)
& \(14.63 \pm 5.95\)
& 28/2/0
& \(1.990{\times}10^{-6}\)
& \(4.693{\times}10^{-5}\)
\\

\(m=5\)
& \(11.52 \pm 6.27\)
& 28/2/0
& \(1.988{\times}10^{-6}\)
& \(4.693{\times}10^{-5}\)
& \(13.93 \pm 6.45\)
& 28/2/0
& \(1.999{\times}10^{-6}\)
& \(4.693{\times}10^{-5}\)
\\

\(m=6\)
& \(12.20 \pm 6.01\)
& 28/2/0
& \(1.991{\times}10^{-6}\)
& \(4.693{\times}10^{-5}\)
& \(13.53 \pm 5.21\)
& 28/2/0
& \(1.991{\times}10^{-6}\)
& \(4.693{\times}10^{-5}\)
\\

\(m=7\)
& \(14.22 \pm 5.98\)
& 28/2/0
& \(1.983{\times}10^{-6}\)
& \(4.693{\times}10^{-5}\)
& \(13.45 \pm 5.81\)
& 28/2/0
& \(1.991{\times}10^{-6}\)
& \(4.693{\times}10^{-5}\)
\\

\(m=8\)
& \(\underline{10.42 \pm 6.01}\)
& 28/2/0
& \(1.974{\times}10^{-6}\)
& \(4.693{\times}10^{-5}\)
& \(13.20 \pm 6.18\)
& 28/2/0
& \(1.983{\times}10^{-6}\)
& \(4.693{\times}10^{-5}\)
\\

\(m=9\)
& \(14.97 \pm 7.15\)
& 28/2/0
& \(1.989{\times}10^{-6}\)
& \(4.693{\times}10^{-5}\)
& \(14.87 \pm 5.38\)
& 28/2/0
& \(1.989{\times}10^{-6}\)
& \(4.693{\times}10^{-5}\)
\\

\(m=10\)
& \(12.90 \pm 6.89\)
& 28/2/0
& \(1.988{\times}10^{-6}\)
& \(4.693{\times}10^{-5}\)
& \(12.98 \pm 5.06\)
& 28/2/0
& \(1.993{\times}10^{-6}\)
& \(4.693{\times}10^{-5}\)
\\

\(m=11\)
& \(12.65 \pm 6.01\)
& 28/2/0
& \(1.994{\times}10^{-6}\)
& \(4.693{\times}10^{-5}\)
& \(14.32 \pm 5.60\)
& 28/2/0
& \(1.989{\times}10^{-6}\)
& \(4.693{\times}10^{-5}\)
\\

\(m=12\)
& \(12.52 \pm 6.70\)
& 28/2/0
& \(1.986{\times}10^{-6}\)
& \(4.693{\times}10^{-5}\)
& \(\underline{12.52 \pm 5.87}\)
& 28/2/0
& \(1.991{\times}10^{-6}\)
& \(4.693{\times}10^{-5}\)
\\

\bottomrule
\end{tabular*}
\end{table}

\begin{table}[H]
\centering
\caption{
Complete capacity-allocation results on the 15 TALENT-tiny regression datasets.
Rank entries report the mean and across-dataset standard deviation within the fixed 25-variant pool. Lower is better.
Underlining marks the best fixed capacity within each geometry.
}
\label{tab:capacity_regression_results}
\scriptsize
\setlength{\tabcolsep}{3.0pt}
\renewcommand{\arraystretch}{1.08}
\begin{tabular*}{\textwidth}{
@{\extracolsep{\fill}}
c
cccc
cccc
@{}
}
\toprule
&
\multicolumn{4}{c}{Hyperbolic HTNN}
&
\multicolumn{4}{c}{HTNN-Euc}
\\
\cmidrule(lr){2-5}
\cmidrule(lr){6-9}
Capacity
& Mean rank
& W/T/L
& Raw \(p\)
& Holm \(p\)
& Mean rank
& W/T/L
& Raw \(p\)
& Holm \(p\)
\\
\midrule
\textbf{Adaptive \(m_j\)}
& \(\mathbf{1.00 \pm 0.00}\)
& ---
& ---
& ---
& \multicolumn{4}{c}{---}
\\
\midrule

\(m=1\)
& \(17.13 \pm 7.04\)
& 15/0/0
& \(6.517{\times}10^{-4}\)
& \(1.506{\times}10^{-2}\)
& \(18.93 \pm 6.10\)
& 15/0/0
& \(6.419{\times}10^{-4}\)
& \(1.506{\times}10^{-2}\)
\\

\(m=2\)
& \(11.13 \pm 5.59\)
& 15/0/0
& \(6.467{\times}10^{-4}\)
& \(1.506{\times}10^{-2}\)
& \(18.07 \pm 6.83\)
& 15/0/0
& \(6.402{\times}10^{-4}\)
& \(1.506{\times}10^{-2}\)
\\

\(m=3\)
& \(10.27 \pm 6.28\)
& 15/0/0
& \(6.484{\times}10^{-4}\)
& \(1.506{\times}10^{-2}\)
& \(15.00 \pm 5.82\)
& 15/0/0
& \(6.500{\times}10^{-4}\)
& \(1.506{\times}10^{-2}\)
\\

\(m=4\)
& \(10.47 \pm 7.00\)
& 15/0/0
& \(6.517{\times}10^{-4}\)
& \(1.506{\times}10^{-2}\)
& \(16.33 \pm 6.11\)
& 15/0/0
& \(6.500{\times}10^{-4}\)
& \(1.506{\times}10^{-2}\)
\\

\(m=5\)
& \(12.00 \pm 7.91\)
& 15/0/0
& \(6.451{\times}10^{-4}\)
& \(1.506{\times}10^{-2}\)
& \(15.27 \pm 7.57\)
& 15/0/0
& \(6.451{\times}10^{-4}\)
& \(1.506{\times}10^{-2}\)
\\

\(m=6\)
& \(9.67 \pm 5.18\)
& 15/0/0
& \(6.435{\times}10^{-4}\)
& \(1.506{\times}10^{-2}\)
& \(17.87 \pm 6.42\)
& 15/0/0
& \(6.386{\times}10^{-4}\)
& \(1.506{\times}10^{-2}\)
\\

\(m=7\)
& \(10.80 \pm 6.32\)
& 15/0/0
& \(6.451{\times}10^{-4}\)
& \(1.506{\times}10^{-2}\)
& \(15.40 \pm 5.78\)
& 15/0/0
& \(6.517{\times}10^{-4}\)
& \(1.506{\times}10^{-2}\)
\\

\(m=8\)
& \(11.60 \pm 7.68\)
& 15/0/0
& \(6.484{\times}10^{-4}\)
& \(1.506{\times}10^{-2}\)
& \(14.40 \pm 6.37\)
& 15/0/0
& \(6.435{\times}10^{-4}\)
& \(1.506{\times}10^{-2}\)
\\

\(m=9\)
& \(\underline{9.13 \pm 5.41}\)
& 15/0/0
& \(6.435{\times}10^{-4}\)
& \(1.506{\times}10^{-2}\)
& \(16.20 \pm 5.85\)
& 15/0/0
& \(6.273{\times}10^{-4}\)
& \(1.506{\times}10^{-2}\)
\\

\(m=10\)
& \(12.00 \pm 6.92\)
& 15/0/0
& \(6.500{\times}10^{-4}\)
& \(1.506{\times}10^{-2}\)
& \(14.13 \pm 6.24\)
& 15/0/0
& \(6.467{\times}10^{-4}\)
& \(1.506{\times}10^{-2}\)
\\

\(m=11\)
& \(9.20 \pm 5.31\)
& 15/0/0
& \(6.354{\times}10^{-4}\)
& \(1.506{\times}10^{-2}\)
& \(14.13 \pm 7.25\)
& 15/0/0
& \(6.517{\times}10^{-4}\)
& \(1.506{\times}10^{-2}\)
\\

\(m=12\)
& \(11.47 \pm 7.69\)
& 15/0/0
& \(6.500{\times}10^{-4}\)
& \(1.506{\times}10^{-2}\)
& \(\underline{13.40 \pm 6.05}\)
& 15/0/0
& \(6.451{\times}10^{-4}\)
& \(1.506{\times}10^{-2}\)
\\

\bottomrule
\end{tabular*}
\end{table}

\section{Per-Dataset Efficiency Results}
\label{app:efficiency_results}

Tables~\ref{tab:efficiency_classification_results} and~\ref{tab:efficiency_regression_results} report the per-dataset timing results for HTNN and its geometry-free inference form.
Fitting times are reported in seconds per 1K training rows, while inference times are reported in seconds per 1K test rows.
Each entry gives the mean and standard deviation over 15 runs.
Because the geometry-free form is obtained through exact post-training conversion, the fitting-time column corresponds to the original HTNN training process.

For dataset \(d\), inference speedup is computed from the normalized mean times defined in Appendix~\ref{app:efficiency_protocol}:
\begin{equation}
s_d
=
\frac{
\bar{t}^{\mathrm{inf}}_{\mathrm{HTNN},d}
}{
\bar{t}^{\mathrm{inf}}_{\mathrm{GF},d}
},
\end{equation}
where \(\mathrm{GF}\) denotes the geometry-free inference form.
Thus, \(s_d>1\) indicates faster inference after conversion.
The task-level speedup is the median of \(s_d\) over the corresponding datasets.

\begin{table}[H]
\centering
\caption{
Per-dataset normalized timing results on the 30 TALENT-tiny
classification datasets.
The speedup column reports original HTNN inference time divided by
geometry-free inference time.
}
\label{tab:efficiency_classification_results}
\scriptsize
\setlength{\tabcolsep}{4.5pt}
\renewcommand{\arraystretch}{1.06}
\begin{tabularx}{\textwidth}{
@{}
>{\raggedright\arraybackslash}X
cccc
@{}
}
\toprule
Dataset
& \shortstack{Fitting\\time}
& \shortstack{HTNN\\inference}
& \shortstack{Geometry-free\\inference}
& Speedup \\
\midrule
ada
& \(48.979 \pm 4.255\)
& \(0.943 \pm 0.079\)
& \(0.590 \pm 0.069\)
& \(1.60\times\) \\

airlines\_2000
& \(4.383 \pm 1.133\)
& \(0.330 \pm 0.047\)
& \(0.177 \pm 0.073\)
& \(1.87\times\) \\

allbp
& \(46.807 \pm 17.763\)
& \(0.582 \pm 0.096\)
& \(0.294 \pm 0.013\)
& \(1.98\times\) \\

ASP--POTASSCO
& \(453.495 \pm 144.814\)
& \(0.985 \pm 0.054\)
& \(0.608 \pm 0.066\)
& \(1.62\times\) \\

autoUniv-au7-1100
& \(70.000 \pm 46.418\)
& \(0.314 \pm 0.026\)
& \(0.146 \pm 0.012\)
& \(2.15\times\) \\

company\_bankruptcy\_prediction
& \(66.614 \pm 41.057\)
& \(1.901 \pm 0.084\)
& \(1.178 \pm 0.100\)
& \(1.61\times\) \\

eucalyptus
& \(114.876 \pm 38.221\)
& \(0.286 \pm 0.017\)
& \(0.121 \pm 0.011\)
& \(2.37\times\) \\

Gender\_Gap\_in\_Spanish\_WP
& \(36.192 \pm 17.235\)
& \(0.440 \pm 0.103\)
& \(0.280 \pm 0.075\)
& \(1.57\times\) \\

hill-valley
& \(657.055 \pm 51.084\)
& \(0.747 \pm 0.148\)
& \(0.471 \pm 0.079\)
& \(1.59\times\) \\

house\_16H
& \(28.802 \pm 6.356\)
& \(0.548 \pm 0.098\)
& \(0.267 \pm 0.060\)
& \(2.05\times\) \\

ibm-employee-performance
& \(86.889 \pm 28.854\)
& \(0.431 \pm 0.063\)
& \(0.179 \pm 0.011\)
& \(2.40\times\) \\

INNHotelsGroup
& \(15.809 \pm 5.254\)
& \(0.877 \pm 0.100\)
& \(0.625 \pm 0.022\)
& \(1.40\times\) \\

internet\_firewall
& \(1.271 \pm 0.818\)
& \(0.772 \pm 0.103\)
& \(0.382 \pm 0.016\)
& \(2.02\times\) \\

jasmine
& \(54.978 \pm 34.532\)
& \(1.717 \pm 0.102\)
& \(1.255 \pm 0.171\)
& \(1.37\times\) \\

jungle\_chess\_2pcs\_raw\_endgame
& \(13.481 \pm 1.297\)
& \(0.514 \pm 0.098\)
& \(0.309 \pm 0.024\)
& \(1.66\times\) \\

law-school-admission-binary
& \(0.268 \pm 0.055\)
& \(0.476 \pm 0.071\)
& \(0.258 \pm 0.018\)
& \(1.85\times\) \\

microaggregation2
& \(25.463 \pm 7.677\)
& \(0.607 \pm 0.106\)
& \(0.311 \pm 0.015\)
& \(1.95\times\) \\

national-longitudinal-survey-binary
& \(19.642 \pm 7.996\)
& \(0.508 \pm 0.104\)
& \(0.228 \pm 0.012\)
& \(2.23\times\) \\

okcupid\_stem
& \(2.920 \pm 1.657\)
& \(1.117 \pm 0.585\)
& \(0.345 \pm 0.028\)
& \(3.24\times\) \\

online\_shoppers
& \(3.395 \pm 0.661\)
& \(0.519 \pm 0.103\)
& \(0.244 \pm 0.018\)
& \(2.12\times\) \\

ozone\_level
& \(0.647 \pm 0.539\)
& \(0.558 \pm 0.107\)
& \(0.413 \pm 0.014\)
& \(1.35\times\) \\

pc4
& \(131.507 \pm 36.235\)
& \(0.470 \pm 0.085\)
& \(0.252 \pm 0.067\)
& \(1.86\times\) \\

PhishingWebsites
& \(37.007 \pm 9.148\)
& \(0.771 \pm 0.104\)
& \(0.548 \pm 0.024\)
& \(1.41\times\) \\

rice\_cammeo\_and\_osmancik
& \(9.203 \pm 6.055\)
& \(0.360 \pm 0.068\)
& \(0.175 \pm 0.014\)
& \(2.05\times\) \\

shill-bidding
& \(7.787 \pm 2.359\)
& \(0.249 \pm 0.012\)
& \(0.137 \pm 0.007\)
& \(1.82\times\) \\

Shipping
& \(11.502 \pm 5.456\)
& \(0.543 \pm 0.098\)
& \(0.368 \pm 0.015\)
& \(1.47\times\) \\

statlog
& \(45.063 \pm 30.596\)
& \(0.330 \pm 0.016\)
& \(0.139 \pm 0.009\)
& \(2.38\times\) \\

thyroid
& \(26.130 \pm 12.894\)
& \(0.670 \pm 0.099\)
& \(0.299 \pm 0.018\)
& \(2.24\times\) \\

waveform\_v1
& \(15.266 \pm 3.867\)
& \(0.561 \pm 0.101\)
& \(0.423 \pm 0.015\)
& \(1.33\times\) \\

wine-quality-red
& \(71.199 \pm 5.311\)
& \(0.254 \pm 0.016\)
& \(0.104 \pm 0.007\)
& \(2.44\times\) \\
\midrule
\multicolumn{4}{r}{Median per-dataset speedup}
& \(\mathbf{1.87\times}\) \\
\bottomrule
\end{tabularx}
\end{table}

\begin{table}[H]
\centering
\caption{
Per-dataset normalized timing results on the 15 TALENT-tiny
regression datasets.
The speedup column reports original HTNN inference time divided by
geometry-free inference time.
}
\label{tab:efficiency_regression_results}
\scriptsize
\setlength{\tabcolsep}{4.5pt}
\renewcommand{\arraystretch}{1.06}
\begin{tabularx}{\textwidth}{
@{}
>{\raggedright\arraybackslash}X
cccc
@{}
}
\toprule
Dataset
& \shortstack{Fitting\\time}
& \shortstack{HTNN\\inference}
& \shortstack{Geometry-free\\inference}
& Speedup \\
\midrule
2dplanes
& \(16.136 \pm 2.076\)
& \(0.640 \pm 0.104\)
& \(0.448 \pm 0.072\)
& \(1.43\times\) \\

Abalone\_reg
& \(52.300 \pm 3.676\)
& \(0.333 \pm 0.013\)
& \(0.316 \pm 0.040\)
& \(1.05\times\) \\

analcatdata\_supreme
& \(42.386 \pm 16.687\)
& \(0.429 \pm 0.090\)
& \(0.272 \pm 0.071\)
& \(1.58\times\) \\

BNG(echoMonths)
& \(10.705 \pm 1.914\)
& \(0.451 \pm 0.078\)
& \(0.352 \pm 0.068\)
& \(1.28\times\) \\

BNG(mv)
& \(7.509 \pm 1.095\)
& \(1.085 \pm 0.096\)
& \(0.688 \pm 0.050\)
& \(1.58\times\) \\

concrete\_compressive\_strength
& \(230.254 \pm 106.320\)
& \(0.425 \pm 0.323\)
& \(0.137 \pm 0.026\)
& \(3.10\times\) \\

cpu\_act
& \(51.010 \pm 8.633\)
& \(0.538 \pm 0.075\)
& \(0.293 \pm 0.044\)
& \(1.84\times\) \\

fifa
& \(13.012 \pm 2.706\)
& \(0.433 \pm 0.108\)
& \(0.189 \pm 0.022\)
& \(2.29\times\) \\

Goodreads-Computer-Books
& \(44.731 \pm 38.870\)
& \(0.845 \pm 0.843\)
& \(0.131 \pm 0.014\)
& \(6.46\times\) \\

house\_8L
& \(6.947 \pm 1.462\)
& \(0.508 \pm 0.099\)
& \(0.267 \pm 0.051\)
& \(1.91\times\) \\

IEEE80211aa-GATS
& \(109.513 \pm 37.563\)
& \(0.954 \pm 0.697\)
& \(0.443 \pm 0.014\)
& \(2.15\times\) \\

Physicochemical\_r
& \(16.205 \pm 1.484\)
& \(0.720 \pm 0.096\)
& \(0.421 \pm 0.073\)
& \(1.71\times\) \\

pole
& \(17.056 \pm 5.825\)
& \(0.740 \pm 0.100\)
& \(0.396 \pm 0.071\)
& \(1.87\times\) \\

qsar\_aquatic\_toxicity
& \(183.522 \pm 38.294\)
& \(0.256 \pm 0.015\)
& \(0.122 \pm 0.016\)
& \(2.09\times\) \\

RSSI\_Estimation
& \(37.095 \pm 9.084\)
& \(0.365 \pm 0.070\)
& \(0.223 \pm 0.062\)
& \(1.63\times\) \\
\midrule
\multicolumn{4}{r}{Median per-dataset speedup}
& \(\mathbf{1.84\times}\) \\
\bottomrule
\end{tabularx}
\end{table}

\end{document}

%% file: math_commands.tex
\usepackage{amsmath,amsfonts,bm}

\def\eqref#1{equation~\ref{#1}}

\def\1{\bm{1}}

\def\ra{{\textnormal{a}}}

\def\rx{{\textnormal{x}}}

\def\rva{{\mathbf{a}}}

\def\erva{{\textnormal{a}}}

\def\ervx{{\textnormal{x}}}

\def\rmA{{\mathbf{A}}}

\def\vmu{{\bm{\mu}}}
\def\vtheta{{\bm{\theta}}}
\def\va{{\bm{a}}}

\def\ve{{\bm{e}}}

\def\vx{{\bm{x}}}

\def\eva{{a}}

\def\mA{{\bm{A}}}

\def\mH{{\bm{H}}}
\def\mI{{\bm{I}}}
\def\mJ{{\bm{J}}}

\def\mX{{\bm{X}}}

\def\mSigma{{\bm{\Sigma}}}

\DeclareMathAlphabet{\mathsfit}{\encodingdefault}{\sfdefault}{m}{sl}
\SetMathAlphabet{\mathsfit}{bold}{\encodingdefault}{\sfdefault}{bx}{n}
\newcommand{\tens}[1]{\bm{\mathsfit{#1}}}
\def\tA{{\tens{A}}}

\def\tX{{\tens{X}}}

\def\gG{{\mathcal{G}}}

\def\sA{{\mathbb{A}}}
\def\sB{{\mathbb{B}}}

\def\sS{{\mathbb{S}}}

\def\emA{{A}}

\newcommand{\etens}[1]{\mathsfit{#1}}

\def\etA{{\etens{A}}}

\newcommand{\E}{\mathbb{E}}

\newcommand{\R}{\mathbb{R}}

\newcommand{\KL}{D_{\mathrm{KL}}}
\newcommand{\Var}{\mathrm{Var}}

\newcommand{\Cov}{\mathrm{Cov}}

\newcommand{\normltwo}{L^2}
\newcommand{\normlp}{L^p}

\newcommand{\parents}{Pa} 

%% file: tables/htnn_per_dataset_results.tex
\begingroup
\scriptsize
\setlength{\tabcolsep}{2pt}
\renewcommand{\arraystretch}{0.98}
\setlength{\LTleft}{0pt}
\setlength{\LTright}{0pt}
\setlength{\LTpre}{4pt}
\setlength{\LTpost}{6pt}
\begin{longtable}{@{}>{\raggedright\arraybackslash}p{0.270\textwidth}>{\raggedleft\arraybackslash}p{0.200\textwidth}@{\hspace{5pt}\vrule width 0.3pt\hspace{5pt}}>{\raggedright\arraybackslash}p{0.270\textwidth}>{\raggedleft\arraybackslash}p{0.200\textwidth}@{}}
\caption{Per-dataset test results for HTNN on the full TALENT benchmark. Classification entries report accuracy (\%), and regression entries report RMSE. Values are means $\pm$ standard deviations over 15 runs.}
\label{tab:htnn_per_dataset_results}\\
\toprule
\textbf{Dataset} & \textbf{Mean $\boldsymbol{\pm}$ Std} & \textbf{Dataset} & \textbf{Mean $\boldsymbol{\pm}$ Std} \\
\midrule
\endfirsthead
\multicolumn{4}{l}{\textit{Table~\ref{tab:htnn_per_dataset_results} continued.}}\\
\toprule
\textbf{Dataset} & \textbf{Mean $\boldsymbol{\pm}$ Std} & \textbf{Dataset} & \textbf{Mean $\boldsymbol{\pm}$ Std} \\
\midrule
\endhead
\midrule
\multicolumn{4}{r}{\textit{Continued on the next page.}}\\
\endfoot
\bottomrule
\endlastfoot
\multicolumn{4}{c}{\textbf{Classification (200 datasets)}}\\
\addlinespace[1pt]
abalone & \(64.68 \pm 0.21\) & KDD & \(79.61 \pm 0.19\) \\
accelerometer & \(74.44 \pm 0.07\) & kdd\_\allowbreak{}ipums\_\allowbreak{}la\_\allowbreak{}97-\allowbreak{}small & \(86.45 \pm 0.19\) \\
ada & \(86.71 \pm 0.14\) & KDDCup09\_\allowbreak{}upselling & \(81.41 \pm 0.15\) \\
ada\_\allowbreak{}agnostic & \(84.69 \pm 0.15\) & kr-\allowbreak{}vs-\allowbreak{}k & \(89.64 \pm 0.15\) \\
ada\_\allowbreak{}prior & \(85.63 \pm 0.13\) & kr-\allowbreak{}vs-\allowbreak{}kp & \(99.75 \pm 0.13\) \\
adult & \(86.09 \pm 0.02\) & kropt & \(91.80 \pm 0.08\) \\
airlines\_\allowbreak{}2000 & \(67.25 \pm 0.42\) & law-\allowbreak{}school-\allowbreak{}admission-\allowbreak{}binary & \(100.00 \pm 0.00\) \\
allbp & \(98.03 \pm 0.04\) & led24 & \(75.00 \pm 0.23\) \\
allrep & \(98.41 \pm 0.10\) & led7 & \(75.12 \pm 0.09\) \\
Amazon\_\allowbreak{}employee\_\allowbreak{}access & \(94.64 \pm 0.02\) & letter & \(98.19 \pm 0.04\) \\
analcatdata\_\allowbreak{}authorship & \(99.64 \pm 0.29\) & Long & \(99.93 \pm 0.05\) \\
artificial-\allowbreak{}characters & \(75.18 \pm 0.44\) & madeline & \(65.41 \pm 0.45\) \\
ASP-\allowbreak{}POTASSCO-\allowbreak{}classification & \(44.59 \pm 0.40\) & MagicTelescope & \(88.51 \pm 0.05\) \\
autoUniv-\allowbreak{}au4-\allowbreak{}2500 & \(63.26 \pm 0.66\) & mammography & \(99.09 \pm 0.02\) \\
autoUniv-\allowbreak{}au7-\allowbreak{}1100 & \(40.05 \pm 0.43\) & Marketing\_\allowbreak{}Campaign & \(89.96 \pm 0.37\) \\
bank & \(91.04 \pm 0.04\) & maternal\_\allowbreak{}health\_\allowbreak{}risk & \(83.60 \pm 0.66\) \\
Bank\_\allowbreak{}Customer\_\allowbreak{}Churn\_\allowbreak{}Dataset & \(87.90 \pm 0.05\) & mfeat-\allowbreak{}factors & \(97.30 \pm 0.10\) \\
banknote\_\allowbreak{}authentication & \(57.13 \pm 0.68\) & mfeat-\allowbreak{}fourier & \(89.75 \pm 0.39\) \\
baseball & \(95.56 \pm 0.20\) & mfeat-\allowbreak{}karhunen & \(96.77 \pm 0.08\) \\
Basketball\_\allowbreak{}c & \(73.28 \pm 0.25\) & mfeat-\allowbreak{}morphological & \(79.00 \pm 0.16\) \\
BLE\_\allowbreak{}RSSI\_\allowbreak{}localization & \(74.51 \pm 0.11\) & mfeat-\allowbreak{}pixel & \(97.40 \pm 0.17\) \\
BNG(breast-\allowbreak{}w) & \(98.63 \pm 0.03\) & mfeat-\allowbreak{}zernike & \(87.88 \pm 0.54\) \\
BNG(cmc) & \(58.62 \pm 0.03\) & MIC & \(85.03 \pm 0.77\) \\
BNG(tic-\allowbreak{}tac-\allowbreak{}toe) & \(82.06 \pm 0.04\) & mice\_\allowbreak{}protein\_\allowbreak{}expression & \(100.00 \pm 0.00\) \\
California-\allowbreak{}Housing-\allowbreak{}Classification & \(89.32 \pm 0.06\) & microaggregation2 & \(64.07 \pm 0.06\) \\
car-\allowbreak{}evaluation & \(100.00 \pm 0.00\) & mobile\_\allowbreak{}c36\_\allowbreak{}oversampling & \(98.51 \pm 0.03\) \\
Cardiovascular-\allowbreak{}Disease-\allowbreak{}dataset & \(73.56 \pm 0.04\) & Mobile\_\allowbreak{}Price\_\allowbreak{}Classification & \(97.68 \pm 0.22\) \\
churn & \(96.28 \pm 0.12\) & mozilla4 & \(93.11 \pm 0.08\) \\
Click\_\allowbreak{}prediction\_\allowbreak{}small & \(83.45 \pm 0.00\) & naticusdroid+\allowbreak{}android+\allowbreak{}permissions & \(97.16 \pm 0.02\) \\
cmc & \(61.46 \pm 0.43\) & national-\allowbreak{}longitudinal-\allowbreak{}survey-\allowbreak{}binary & \(99.69 \pm 0.00\) \\
company\_\allowbreak{}bankruptcy\_\allowbreak{}prediction & \(97.38 \pm 0.03\) & Nutrition\_\allowbreak{}Health\_\allowbreak{}Survey & \(85.37 \pm 0.28\) \\
compass & \(78.11 \pm 0.24\) & okcupid\_\allowbreak{}stem & \(75.52 \pm 0.08\) \\
connect-\allowbreak{}4 & \(86.08 \pm 0.10\) & one-\allowbreak{}hundred-\allowbreak{}plants-\allowbreak{}margin & \(85.41 \pm 0.31\) \\
Contaminant-\allowbreak{}10.0GHz & \(95.63 \pm 0.09\) & one-\allowbreak{}hundred-\allowbreak{}plants-\allowbreak{}shape & \(74.34 \pm 0.57\) \\
Contaminant-\allowbreak{}10.5GHz & \(95.83 \pm 0.21\) & one-\allowbreak{}hundred-\allowbreak{}plants-\allowbreak{}texture & \(89.38 \pm 0.40\) \\
Contaminant-\allowbreak{}11.0GHz & \(95.10 \pm 0.21\) & online\_\allowbreak{}shoppers & \(90.82 \pm 0.13\) \\
Contaminant-\allowbreak{}9.0GHz & \(93.27 \pm 0.28\) & optdigits & \(99.08 \pm 0.08\) \\
Contaminant-\allowbreak{}9.5GHz & \(93.65 \pm 0.14\) & ozone-\allowbreak{}level-\allowbreak{}8hr & \(96.00 \pm 0.09\) \\
contraceptive\_\allowbreak{}method\_\allowbreak{}choice & \(63.69 \pm 0.39\) & ozone\_\allowbreak{}level & \(98.23 \pm 0.00\) \\
credit & \(76.52 \pm 0.06\) & page-\allowbreak{}blocks & \(97.48 \pm 0.06\) \\
credit-\allowbreak{}g & \(81.05 \pm 0.57\) & pc1 & \(94.64 \pm 0.14\) \\
Credit\_\allowbreak{}c & \(77.25 \pm 0.06\) & pc3 & \(90.22 \pm 0.46\) \\
Customer\_\allowbreak{}Personality\_\allowbreak{}Analysis & \(90.42 \pm 0.35\) & pc4 & \(91.71 \pm 0.34\) \\
customer\_\allowbreak{}satisfaction\_\allowbreak{}in\_\allowbreak{}airline & \(96.18 \pm 0.02\) & pendigits & \(99.70 \pm 0.02\) \\
dabetes\_\allowbreak{}130-\allowbreak{}us\_\allowbreak{}hospitals & \(64.52 \pm 0.08\) & Performance-\allowbreak{}Prediction & \(76.04 \pm 0.44\) \\
default\_\allowbreak{}of\_\allowbreak{}credit\_\allowbreak{}card\_\allowbreak{}clients & \(82.82 \pm 0.04\) & philippine & \(73.29 \pm 0.20\) \\
delta\_\allowbreak{}ailerons & \(95.06 \pm 0.04\) & PhishingWebsites & \(97.88 \pm 0.06\) \\
Diabetic\_\allowbreak{}Retinopathy\_\allowbreak{}Debrecen & \(79.78 \pm 0.48\) & phoneme & \(89.74 \pm 0.21\) \\
dis & \(98.98 \pm 0.06\) & PieChart3 & \(89.40 \pm 0.60\) \\
dna & \(96.30 \pm 0.08\) & Pima\_\allowbreak{}Indians\_\allowbreak{}Diabetes\_\allowbreak{}Database & \(81.82 \pm 0.92\) \\
drug\_\allowbreak{}consumption & \(40.72 \pm 0.32\) & PizzaCutter3 & \(90.24 \pm 0.38\) \\
dry\_\allowbreak{}bean\_\allowbreak{}dataset & \(93.15 \pm 0.03\) & pol & \(99.37 \pm 0.04\) \\
E-\allowbreak{}CommereShippingData & \(67.40 \pm 0.29\) & predict\_\allowbreak{}students\_\allowbreak{}dropout & \(76.89 \pm 0.17\) \\
eeg-\allowbreak{}eye-\allowbreak{}state & \(97.21 \pm 0.22\) & Pumpkin\_\allowbreak{}Seeds & \(87.70 \pm 0.13\) \\
electricity & \(87.21 \pm 0.13\) & QSAR\_\allowbreak{}biodegradation & \(91.28 \pm 0.38\) \\
Employee & \(86.36 \pm 0.11\) & Rain\_\allowbreak{}in\_\allowbreak{}Australia & \(86.36 \pm 0.03\) \\
estimation\_\allowbreak{}of\_\allowbreak{}obesity\_\allowbreak{}levels & \(98.56 \pm 0.20\) & rice\_\allowbreak{}cammeo\_\allowbreak{}and\_\allowbreak{}osmancik & \(93.01 \pm 0.08\) \\
eucalyptus & \(77.03 \pm 0.60\) & ringnorm & \(98.22 \pm 0.03\) \\
eye\_\allowbreak{}movements & \(65.58 \pm 0.31\) & rl & \(73.04 \pm 0.46\) \\
eye\_\allowbreak{}movements\_\allowbreak{}bin & \(60.74 \pm 0.17\) & satimage & \(92.37 \pm 0.19\) \\
FICO-\allowbreak{}HELOC-\allowbreak{}cleaned & \(74.41 \pm 0.15\) & SDSS17 & \(97.24 \pm 0.02\) \\
Firm-\allowbreak{}Teacher-\allowbreak{}Direction & \(87.81 \pm 0.13\) & segment & \(93.46 \pm 0.19\) \\
first-\allowbreak{}order-\allowbreak{}theorem-\allowbreak{}proving & \(59.64 \pm 0.20\) & seismic+\allowbreak{}bumps & \(93.44 \pm 0.06\) \\
Fitness\_\allowbreak{}Club\_\allowbreak{}c & \(80.00 \pm 0.39\) & semeion & \(94.11 \pm 0.39\) \\
FOREX\_\allowbreak{}audcad-\allowbreak{}day-\allowbreak{}High & \(76.19 \pm 0.25\) & shill-\allowbreak{}bidding & \(91.15 \pm 0.04\) \\
FOREX\_\allowbreak{}audcad-\allowbreak{}hour-\allowbreak{}High & \(71.93 \pm 0.07\) & Shipping & \(68.86 \pm 0.16\) \\
FOREX\_\allowbreak{}audchf-\allowbreak{}day-\allowbreak{}High & \(77.36 \pm 0.19\) & shrutime & \(87.26 \pm 0.11\) \\
FOREX\_\allowbreak{}audjpy-\allowbreak{}day-\allowbreak{}High & \(78.39 \pm 0.42\) & shuttle & \(99.99 \pm 0.00\) \\
FOREX\_\allowbreak{}audjpy-\allowbreak{}hour-\allowbreak{}High & \(71.18 \pm 0.15\) & spambase & \(94.77 \pm 0.08\) \\
FOREX\_\allowbreak{}audsgd-\allowbreak{}hour-\allowbreak{}High & \(69.13 \pm 0.18\) & splice & \(95.41 \pm 0.23\) \\
FOREX\_\allowbreak{}audusd-\allowbreak{}hour-\allowbreak{}High & \(67.76 \pm 0.17\) & sports\_\allowbreak{}articles & \(85.95 \pm 0.35\) \\
FOREX\_\allowbreak{}cadjpy-\allowbreak{}day-\allowbreak{}High & \(75.12 \pm 0.59\) & statlog & \(75.55 \pm 0.47\) \\
FOREX\_\allowbreak{}cadjpy-\allowbreak{}hour-\allowbreak{}High & \(69.65 \pm 0.20\) & steel\_\allowbreak{}plates\_\allowbreak{}faults & \(79.85 \pm 0.49\) \\
GAMETES\_\allowbreak{}Epistasis & \(66.06 \pm 0.88\) & svmguide3 & \(85.46 \pm 0.30\) \\
GAMETES\_\allowbreak{}Heterogeneity & \(70.03 \pm 0.94\) & sylvine & \(95.38 \pm 0.21\) \\
gas-\allowbreak{}drift & \(99.50 \pm 0.02\) & taiwanese\_\allowbreak{}bankruptcy\_\allowbreak{}prediction & \(97.20 \pm 0.06\) \\
Gender\_\allowbreak{}Gap\_\allowbreak{}in\_\allowbreak{}Spanish\_\allowbreak{}WP & \(61.89 \pm 0.22\) & telco-\allowbreak{}customer-\allowbreak{}churn & \(80.59 \pm 0.07\) \\
GesturePhaseSegmentationProcessed & \(68.28 \pm 0.19\) & Telecom\_\allowbreak{}Churn\_\allowbreak{}Dataset & \(94.38 \pm 0.12\) \\
golf\_\allowbreak{}play\_\allowbreak{}dataset\_\allowbreak{}extended & \(95.80 \pm 0.49\) & texture & \(99.91 \pm 0.00\) \\
Heart-\allowbreak{}Disease-\allowbreak{}Dataset & \(92.98 \pm 0.38\) & thyroid & \(99.39 \pm 0.04\) \\
helena & \(38.90 \pm 0.06\) & thyroid-\allowbreak{}ann & \(99.11 \pm 0.06\) \\
heloc & \(73.13 \pm 0.09\) & thyroid-\allowbreak{}dis & \(71.27 \pm 0.35\) \\
hill-\allowbreak{}valley & \(79.59 \pm 2.12\) & turiye\_\allowbreak{}student\_\allowbreak{}evaluation & \(53.21 \pm 0.23\) \\
house\_\allowbreak{}16H & \(88.34 \pm 0.07\) & twonorm & \(98.20 \pm 0.05\) \\
HR\_\allowbreak{}Analytics & \(80.34 \pm 0.09\) & UJI\_\allowbreak{}Pen\_\allowbreak{}Characters & \(56.30 \pm 0.40\) \\
htru & \(98.09 \pm 0.02\) & vehicle & \(85.88 \pm 0.64\) \\
ibm-\allowbreak{}employee-\allowbreak{}performance & \(99.83 \pm 0.17\) & VulNoneVul & \(98.96 \pm 0.03\) \\
in\_\allowbreak{}vehicle\_\allowbreak{}coupon & \(77.54 \pm 0.13\) & walking-\allowbreak{}activity & \(64.90 \pm 0.11\) \\
Indian\_\allowbreak{}pines & \(94.66 \pm 0.14\) & wall-\allowbreak{}robot-\allowbreak{}navigation & \(96.79 \pm 0.10\) \\
INNHotelsGroup & \(88.52 \pm 0.14\) & water\_\allowbreak{}quality & \(91.68 \pm 0.15\) \\
Insurance & \(75.77 \pm 0.03\) & Water\_\allowbreak{}Quality\_\allowbreak{}and\_\allowbreak{}Potability & \(66.20 \pm 0.27\) \\
internet\_\allowbreak{}firewall & \(79.79 \pm 0.00\) & Waterstress & \(72.18 \pm 0.56\) \\
internet\_\allowbreak{}usage & \(53.51 \pm 0.17\) & waveform-\allowbreak{}5000 & \(87.53 \pm 0.20\) \\
Intersectional-\allowbreak{}Bias-\allowbreak{}Assessment & \(94.89 \pm 0.05\) & waveform\_\allowbreak{}v1 & \(87.74 \pm 0.15\) \\
Is-\allowbreak{}this-\allowbreak{}a-\allowbreak{}good-\allowbreak{}customer & \(89.97 \pm 0.23\) & website\_\allowbreak{}phishing & \(92.84 \pm 0.24\) \\
JapaneseVowels & \(99.52 \pm 0.03\) & Wilt & \(99.48 \pm 0.00\) \\
jasmine & \(80.55 \pm 0.17\) & wine & \(75.85 \pm 0.35\) \\
jm1 & \(82.05 \pm 0.14\) & wine-\allowbreak{}quality-\allowbreak{}red & \(63.97 \pm 0.52\) \\
jungle\_\allowbreak{}chess\_\allowbreak{}2pcs\_\allowbreak{}raw\_\allowbreak{}endgame & \(98.99 \pm 0.07\) & wine-\allowbreak{}quality-\allowbreak{}white & \(63.65 \pm 0.39\) \\
kc1 & \(88.96 \pm 0.12\) & yeast & \(61.85 \pm 0.48\) \\
\midrule
\multicolumn{4}{c}{\textbf{Regression (100 datasets)}}\\
\addlinespace[1pt]
1000-\allowbreak{}Cameras-\allowbreak{}Dataset & \(462.8 \pm 5.823\) & houses & \(4.87\!\times\!10^{4} \pm 104.4\) \\
2dplanes & \(1.001 \pm 1.13\!\times\!10^{-4}\) & housing\_\allowbreak{}price\_\allowbreak{}prediction & \(9.91\!\times\!10^{5} \pm 6758\) \\
Abalone\_\allowbreak{}reg & \(2.076 \pm 0.001796\) & IEEE80211aa-\allowbreak{}GATS & \(0.01949 \pm 5.34\!\times\!10^{-4}\) \\
Ailerons & \(1.52\!\times\!10^{-4} \pm 3.26\!\times\!10^{-7}\) & Job\_\allowbreak{}Profitability & \(13 \pm 0.0293\) \\
airfoil\_\allowbreak{}self\_\allowbreak{}noise & \(1.229 \pm 0.01859\) & kin8nm & \(0.06502 \pm 3.77\!\times\!10^{-4}\) \\
analcatdata\_\allowbreak{}supreme & \(0.07597 \pm 0.001587\) & Laptop\_\allowbreak{}Prices\_\allowbreak{}Dataset & \(452.4 \pm 2.397\) \\
archive2 & \(357 \pm 5.475\) & mauna-\allowbreak{}loa-\allowbreak{}atmospheric-\allowbreak{}co2 & \(0.4715 \pm 0.007237\) \\
archive\_\allowbreak{}r56\_\allowbreak{}Portuguese & \(2.732 \pm 0.03814\) & MiamiHousing2016 & \(8.59\!\times\!10^{4} \pm 428.5\) \\
auction\_\allowbreak{}verification & \(1991 \pm 102.3\) & MIP-\allowbreak{}2016-\allowbreak{}regression & \(2.4\!\times\!10^{4} \pm 102.7\) \\
avocado\_\allowbreak{}sales & \(0.1072 \pm 4.01\!\times\!10^{-4}\) & Mobile\_\allowbreak{}Phone\_\allowbreak{}Market & \(741.9 \pm 19.13\) \\
bank32nh & \(0.07593 \pm 1.42\!\times\!10^{-4}\) & Moneyball & \(20.16 \pm 0.1555\) \\
bank8FM & \(0.02776 \pm 7.13\!\times\!10^{-5}\) & mv & \(0.005557 \pm 3.63\!\times\!10^{-4}\) \\
Bias\_\allowbreak{}correction\_\allowbreak{}r & \(0.7375 \pm 0.004439\) & NASA\_\allowbreak{}PHM2008 & \(40.17 \pm 0.01303\) \\
Bias\_\allowbreak{}correction\_\allowbreak{}r\_\allowbreak{}2 & \(0.5539 \pm 0.003991\) & NHANES\_\allowbreak{}age\_\allowbreak{}prediction & \(15.36 \pm 0.01155\) \\
bike\_\allowbreak{}sharing\_\allowbreak{}demand & \(70.58 \pm 0.2054\) & OnlineNewsPopularity & \(8572 \pm 1.313\) \\
BNG(echoMonths) & \(11.53 \pm 0.006885\) & Parkinson\_\allowbreak{}Sound\_\allowbreak{}Record & \(14.91 \pm 0.0507\) \\
BNG(lowbwt) & \(456.6 \pm 0.0982\) & Parkinsons\_\allowbreak{}Telemonitoring & \(1.704 \pm 0.1426\) \\
BNG(mv) & \(4.603 \pm 0.002507\) & Physicochemical\_\allowbreak{}r & \(3.369 \pm 0.005011\) \\
BNG(stock) & \(2.95 \pm 0.003955\) & pol\_\allowbreak{}reg & \(2.008 \pm 0.03007\) \\
boston & \(3.862 \pm 0.04451\) & pole & \(1.957 \pm 0.03273\) \\
Brazilian\_\allowbreak{}houses\_\allowbreak{}reproduced & \(0.003602 \pm 5.47\!\times\!10^{-4}\) & puma32H & \(0.005717 \pm 2.49\!\times\!10^{-5}\) \\
chscase\_\allowbreak{}foot & \(0.8221 \pm 0.001891\) & puma8NH & \(3.229 \pm 0.001481\) \\
colleges & \(0.1557 \pm 5.71\!\times\!10^{-4}\) & qsar\_\allowbreak{}aquatic\_\allowbreak{}toxicity & \(0.9467 \pm 0.0103\) \\
combined\_\allowbreak{}cycle\_\allowbreak{}power\_\allowbreak{}plant & \(3.709 \pm 0.02767\) & qsar\_\allowbreak{}fish\_\allowbreak{}toxicity & \(0.8468 \pm 0.005675\) \\
communities\_\allowbreak{}and\_\allowbreak{}crime & \(0.1318 \pm 3.96\!\times\!10^{-4}\) & RSSI\_\allowbreak{}Estimation & \(0.005501 \pm 0.001442\) \\
concrete\_\allowbreak{}compressive\_\allowbreak{}strength & \(4.441 \pm 0.05308\) & RSSI\_\allowbreak{}Estimation1 & \(0.002037 \pm 7.46\!\times\!10^{-4}\) \\
CookbookReviews & \(1.489 \pm 4.08\!\times\!10^{-4}\) & SAT11-\allowbreak{}HAND-\allowbreak{}runtime & \(1062 \pm 3.758\) \\
CPMP-\allowbreak{}2015-\allowbreak{}regression & \(491 \pm 4.679\) & satellite\_\allowbreak{}image & \(0.6004 \pm 0.008006\) \\
CPS1988 & \(363.6 \pm 0.0927\) & sensory & \(0.6752 \pm 0.006116\) \\
cpu\_\allowbreak{}act & \(2.312 \pm 0.01599\) & Shop\_\allowbreak{}Customer\_\allowbreak{}Data & \(28.42 \pm 0.02303\) \\
cpu\_\allowbreak{}small & \(2.845 \pm 0.01152\) & socmob & \(15.93 \pm 2.231\) \\
Data\_\allowbreak{}Science\_\allowbreak{}Salaries & \(5.96\!\times\!10^{4} \pm 91.8\) & space\_\allowbreak{}ga & \(0.09417 \pm 4.25\!\times\!10^{-4}\) \\
dataset\_\allowbreak{}sales & \(3.889 \pm 0.006586\) & steel\_\allowbreak{}industry\_\allowbreak{}energy & \(0.386 \pm 0.04383\) \\
debutanizer & \(0.06088 \pm 0.001625\) & stock & \(0.704 \pm 0.009581\) \\
delta\_\allowbreak{}elevators & \(0.001367 \pm 3.86\!\times\!10^{-7}\) & stock\_\allowbreak{}fardamento02 & \(18.43 \pm 0.05335\) \\
Diamonds & \(564.1 \pm 6.122\) & sulfur & \(0.01796 \pm 5.11\!\times\!10^{-4}\) \\
elevators & \(0.00184 \pm 3.04\!\times\!10^{-6}\) & Superconductivty & \(10.16 \pm 0.04896\) \\
Facebook\_\allowbreak{}Comment\_\allowbreak{}Volume & \(16.58 \pm 0.1995\) & topo\_\allowbreak{}2\_\allowbreak{}1 & \(0.03468 \pm 4.39\!\times\!10^{-6}\) \\
Fiat & \(748.6 \pm 2.073\) & treasury & \(0.2015 \pm 0.00453\) \\
fifa & \(0.7925 \pm 9.99\!\times\!10^{-4}\) & us\_\allowbreak{}crime & \(0.1351 \pm 3.62\!\times\!10^{-4}\) \\
Food\_\allowbreak{}Delivery\_\allowbreak{}Time & \(7.836 \pm 0.002023\) & volume & \(39.62 \pm 0.3411\) \\
fried & \(1.004 \pm 4\!\times\!10^{-4}\) & Wave\_\allowbreak{}Energy\_\allowbreak{}Perth\_\allowbreak{}100 & \(1.68\!\times\!10^{4} \pm 211.7\) \\
garments\_\allowbreak{}worker\_\allowbreak{}productivity & \(0.1475 \pm 0.001885\) & Wave\_\allowbreak{}Energy\_\allowbreak{}Sydney\_\allowbreak{}100 & \(1.93\!\times\!10^{4} \pm 407.1\) \\
gas\_\allowbreak{}turbine\_\allowbreak{}emission & \(0.4401 \pm 0.003682\) & Wave\_\allowbreak{}Energy\_\allowbreak{}Sydney\_\allowbreak{}49 & \(4580 \pm 48.73\) \\
Goodreads-\allowbreak{}Computer-\allowbreak{}Books & \(0.4812 \pm 0.001904\) & weather\_\allowbreak{}izmir & \(1.094 \pm 0.007075\) \\
healthcare\_\allowbreak{}insurance\_\allowbreak{}expenses & \(4678 \pm 18.25\) & wind & \(2.831 \pm 0.00412\) \\
house\_\allowbreak{}16H\_\allowbreak{}reg & \(2.92\!\times\!10^{4} \pm 86.74\) & wine+\allowbreak{}quality & \(0.6947 \pm 0.001761\) \\
house\_\allowbreak{}8L & \(2.8\!\times\!10^{4} \pm 130.4\) & Wine\_\allowbreak{}Quality\_\allowbreak{}red & \(0.6297 \pm 0.005276\) \\
house\_\allowbreak{}prices\_\allowbreak{}nominal & \(2.34\!\times\!10^{4} \pm 536\) & Wine\_\allowbreak{}Quality\_\allowbreak{}white & \(0.6595 \pm 0.002086\) \\
house\_\allowbreak{}sales\_\allowbreak{}reduced & \(1.17\!\times\!10^{5} \pm 424.6\) & yprop\_\allowbreak{}4\_\allowbreak{}1 & \(0.02666 \pm 4.83\!\times\!10^{-5}\) \\
\end{longtable}
\endgroup